%% file: ewrl_2025_mine.tex
\documentclass{article}

\PassOptionsToPackage{numbers, compress}{natbib}

     \usepackage[final]{ewrl_2025}

\usepackage[utf8]{inputenc} 
\usepackage[T1]{fontenc}    
\usepackage{hyperref}       
\usepackage{url}            
\usepackage{booktabs}       
\usepackage{amsfonts}       
\usepackage{nicefrac}       
\usepackage{microtype}      
\usepackage{xcolor}         
\usepackage[capitalise,nameinlink]{cleveref} 
\usepackage{graphicx} 
\usepackage{subcaption} 
\usepackage[disable,prependcaption,textsize=footnotesize]{todonotes}
\usepackage{appendix}
\usepackage{multirow}
\AtBeginEnvironment{appendices}{\crefalias{section}{appendix}}

\title{Exploring a Graph-based Approach to Offline Reinforcement Learning for Sepsis Treatment}

\author{
	Taisiya Khakharova \\
	Chair of Practical Computer Science/Software Systems Engineering \\
	Brandenburg University of Technology Cottbus-Senftenberg, Germany \\	 
	\texttt{Taisiya.Khakharova@b-tu.de} \\
	\AND
	Lucas Sakizloglou \\
	Chair of Practical Computer Science/Software Systems Engineering \\
	Brandenburg University of Technology Cottbus-Senftenberg, Germany \\	 
	\texttt{Lucas.Sakizloglou@b-tu.de} \\
	\AND
	Leen Lambers \\
	Chair of Practical Computer Science/Software Systems Engineering \\
	Brandenburg University of Technology Cottbus-Senftenberg, Germany \\	 
	\texttt{Leen.Lambers@b-tu.de} \\
}

\begin{document}

\maketitle

\begin{abstract}
  Sepsis is a serious, life-threatening condition. When treating sepsis, it is challenging to determine the correct amount of intravenous fluids and vasopressors for a given patient. While automated reinforcement learning (RL)-based methods have been used to support these decisions with promising results, previous studies have relied on relational data. Given the complexity of modern healthcare data, representing data as a graph may provide a more natural and effective approach. This study models patient data from the well-known MIMIC-III dataset as a heterogeneous graph that evolves over time. Subsequently, we explore two Graph Neural Network architectures - GraphSAGE and GATv2 - for learning patient state representations, adopting the approach of decoupling representation learning from policy learning. The encoders are trained to produce latent state representations, jointly with decoders that predict the next patient state. These representations are then used for policy learning with the dBCQ algorithm. The results of our experimental evaluation confirm the potential of a graph-based approach, while highlighting the complexity of representation learning in this domain.
  
\end{abstract}

\input{intro.tex} 
\input{background.tex}
\input{solution.tex}
\input{experiments.tex}
\input{relatedwork.tex}
\input{conclusion.tex}

\bibliographystyle{plain}
\bibliography{sources}

\input{appendix.tex}

\end{document}

%% file: intro.tex
\section{Introduction}
\label{sec:intro}

\todo{Sepsis. Standart way of treatment.}
Sepsis is a severe condition in which the body dysfunctionally responds to infection injuring the body's own tissues and organs, which may escalate to septic shock and lead, consequently, to death. Sepsis is not uncommon: an estimated 15 out of every 1000 hospitalized patients will develop sepsis while sepsis-related deaths account for 20\% of all global deaths \citep{rudd_global_2020}. 

\todo{Traditional approach and an issue indicated}
Vasopressors and intravenous (IV) fluids are essential components in sepsis treatment \cite{waechter_interaction_2014, rhodes_surviving_2017}. IV fluids are used to restore circulating blood volume, while vasopressors address hypotension.

The Surviving Sepsis Campaign guidelines  \cite{rhodes_surviving_2017} proposed a universal approach, suggesting a fixed dosage of the drug per kilogram of patient body weight. However, this approach has been called into question by more recent studies \cite{moschopoulos_new_2023, carlos_sanchez_fluids_2023}, which indicate that personalized treatment strategies should be preferred.

\todo{Sepsis treatment was addresed by RL}
Sepsis treatment has been addressed with reinforcement learning (RL) because it naturally fits a Markov decision process (MDP) framework. Many studies have demonstrated promising results in developing personalized treatment policies that could improve patient outcomes compared to standard protocols \citep{komorowski_artificial_2018, roggeveen_transatlantic_2021, zhang_optimizing_2024, wu_value-based_2023}. RL agents are trained to recommend drug prescriptions based on publicly available medical datasets.
 Specifically, the agents learned to continually recommend a dosage of vasopressors and IV fluids that should be administered to a patient, taking into account multiple clinical variables such as vital signs, laboratory results, and demographic information. 
 
 \todo{Medical data should be encoded as graphs}
 To the best of our knowledge, all the previous studies on RL-based sepsis treatment optimization rely on relational data. As shown by related work \cite{wanyan_deep_2021}, the complexity of medical data requires new architectures for encoding the information, and utilizing graph data structures can positively influence the performance of machine learning models. Graphs can capture semantic relationships for medical data, making further data analysis more efficient and accurate \cite{health_knowledge_graph_aldughayfiq_capturing_2023}.
 
 \todo{GNNs}
 Graph neural network (GNNs) is a type of neural network that learns from graph-structured data. GNNs are designed to capture complex relationships by exploiting structure. Besides being already useful to healthcare research, e.g., for disease prediction \citep{Sharma2023}, recent research from other application domains indicates that GNNs may be more suitable to RL over graph-structured data \citep{drl-and-gnns, football_niu_graph_2022}.

\todo{Problem statement}
In summary, RL  for sepsis treatment is an area of ongoing research that has already shown impressive results. However, the studies are limited to the relational approach: they use relational data and traditional neural networks (NNs), while several studies have highlighted the potential for increased accuracy and efficiency when using medical graph data as input to the machine learning models. This paper investigates how a graph-based approach impacts accuracy and efficiency in RL-optimized sepsis treatment by leveraging GNNs to process the patient graph data.

\todo{Our contributions overview}
In this work, we present a way of modeling timestamped medical data from the dataset MIMIC-III \cite{johnson_mimic-iii_2016} as a graph. Then, we compare two state-of-the-art GNN-based encoder architectures. The GNN-based encoders are used to encode the graph data into latent representations. The RL policies are trained with dBCQ algorithm \cite{fujimoto_benchmarking_2019}. Then, we evaluate the learned RL policies obtained using these latent representations, and compare them with eath other and with the policy learned from representations that were obtained from traditional NN.

The work of Killian et al. \cite{killian_empirical_2020} informs the design of our study. Similar to their design we decouple representation learning from RL policy learning to isolate the effect of the different encoders alone and therefore allow for a fair evaluation. We also compare our results with those of this study.

The rest of this paper is organized as follows: \cref{sec:prelim} explains RL terminology, formulates sepsis treatment as POMDP, describes our data, reviews graphs and GNNs; \cref{sec:approach} presents our approach, with key contributions in \cref{subsec:graph-modeling} and \cref{subsec:gnn-encoder}. \cref{sec:empiricalstudy} provides details of the experimental setup, evaluation and threats to validity;\cref{sec:relatedwork} presents related work; and \cref{sec:conclusion} concludes the paper and outlines future work directions.

%% file: background.tex
\section{Preliminaries}
\label{sec:prelim}

\subsection{Reinforcement Learning}
\todo{RL, offline RL, BCQ, WIS evaluation}

RL \cite{sutton2018reinforcement} is a subfield of machine learning that introduces concepts of agent and environment. The aim for an agent is to learn how to maximize cumulative reward gained through sequential interactions with the environment. Markov Decision Processes (MDP) provide a mathetical framework to describe the environment.

Offline or batch RL refers to a setting when an agent cannot interract with the environment while learning, but only has experiences documented from the past. The medical domain is a safety critical domain, therefore, we are limited to the offline RL application. In this setting, we must prevent the estimated value function from assigning values to actions that do not appear in the provided data. Batch Constrained Q-Learning (BCQ) \cite{fujimoto_off-policy_dqn_ddpg_fails_2018} addresses this problem. For a setting where the action space is discretized there is an adapted version called dBCQ \cite{fujimoto_benchmarking_2019}.

Weighted Importance Sampling (WIS) is one of the fundamental approaches to off-policy evaluation. WIS addresses the distribution mismatch between behaviour and target policies by reweighting trajectories based on their likelihood ratios. The method has been adopted in healthcare applications, including sepsis treatment \cite{killian_empirical_2020, li_optimizing_2019}.

\subsection{Sepsis Treatment as POMDP Problem }
\label{subsec:sepsis-mdp}

We build on the optimization of sepsis treatments based on \textcolor{black}{a partially observable MDP (POMDP)}, as presented in \cite{killian_empirical_2020}. In this work, an RL agent is trained to make personalized vasopressors and IV fluid dosage suggestions based on the patient's vital signs, lab results and demographic information. In further detail, we reuse the following POMDP elements:

\textbf{Action space.}  Dosages of vasopressors and IV fluids were split into five quartiles each resulting into 25 possible actions. Therefore, administration of medication is represented by the discrete value from 0 to 25. \\
\textbf{Reward.} Each patient trajectory is associated with a value of +1 if the patient survived and with a value of -1 if not. The reward is given only once per trajectory - in the end. \\
\textbf{Observation space.} The general idea is that a patient's vital signs, lab values and demographic information are aggregated over 4 hours and encoded into the observation space. We opt for a continuous observation space where each observation is the patient information on a currect time step encoded into a latent vector of the dimesionality $ \hat{d_s} $, as done in \cite{killian_empirical_2020}.

\subsection{Medical Dataset and its Modeling as a Graph}
\todo{mimic-iii and data description}
Similarly to \cite{komorowski_artificial_2018, killian_empirical_2020}, we evaluate our approach based on the MIMIC-III dataset (v.1.4) that contains data from patients in an Intensive Care Unit \cite{johnson_mimic-iii_2016}. Specifically, we reuse the Python implementation of Killian et al.  \cite{killian_empirical_2020} of the MATLAB solution of Komorowski et al.  \cite{komorowski_artificial_2018} for all preprocessing and sepsis cohort extraction, resulting in a cleaned sepsis dataset. The sepsis dataset consists of 18,908 patient trajectories that are independent from each other. Killian et al. additionally remove patient trajectories consisting only of one time step and ended up with 18,573 patient trajectories. The mortality rate in the data remained at approximately 6\%. Each trajectory consists of up to 20 time steps.  Each time step comprises 38 patient features. The features encompass demographic information, lab values and vital signs, which are aggregated over a period of 4 hours. A time step is followed by an administration of a combination of vasopressors and IV fluids.

As mentioned in \cref{sec:intro}, however, contrary to \cite{killian_empirical_2020, komorowski_artificial_2018}, our work explores a graph-based approach. Specifically, we model the dataset as a \emph{dynamic heterogeneous} graph---see \cref{subsec:graph-modeling}: A heterogeneous graph contains multiple types of nodes and/or edges, allowing richer relational structures—e.g., authors, papers, and venues connected by “writes” or “cites” relations \cite{wang_survey_2020}, whereas a dynamic graph changes over time, with nodes or edges appearing, disappearing, or altering attributes over time \cite{Kazemi2022}. On a formal level, such graphs can be represented via suitably defined \textit{typed attributed graphs} \cite{ehrig_hutchison_fundamental_2004}. Typed attributed graphs are graphs whose nodes and edges can have different types, and furthermore be equipped with attributes. Therefore, typed attributed graphs are capable of capturing the information present in the dataset in~\cref{subsec:graph-modeling}. Typed attributed graphs are created based on a \textit{type graph}, that is, an ontology that defines all valid graph instances, i.e., typed attributed graphs.

\subsection{Graph Neural Networks}
\todo{Graph, heterogeneous, dynamic}

Graph Neural Networks (GNNs) \cite{zhou_graph_2021} are deep learning models designed to work with graph-structured data. Unlike traditional neural networks that process fixed-format data (like images or sequences), GNNs can handle irregular structures where each node might have a different number of connections. The key idea behind GNNs is that they learn representations for nodes by aggregating information from their neighbors in the graph.

GNNs update each node’s representation through learnable message-passing. In each layer, a node first aggregates features from its neighbors and then applies a trainable transformation—typically a weighted linear layer followed by an activation function—to produce a new feature vector. Stacking L such layers means each node’s features are updated L times, allowing the GNN to capture progressively a wider graph structure.

%% file: solution.tex
\section{Approach}
\label{sec:approach}

In this section, we provide an overview of our approach and highlight the main contributions: we present a way of modeling medical data a graph and the design of GNN-based encoders.

\subsection{Overview}
\label{subsec:overview}

\todo{Building blocks - general}
The approach consists of the following building blocks: the Data provider, the Encoder, the Decoder, the BCQ policy. We use the Data provider, the Encoder, and the Decoder for representation learning, while the Data provider, the Encoder, and the BCQ policy are used for policy learning.

\begin{figure*}
	\centering	
	\includegraphics[width=\textwidth]{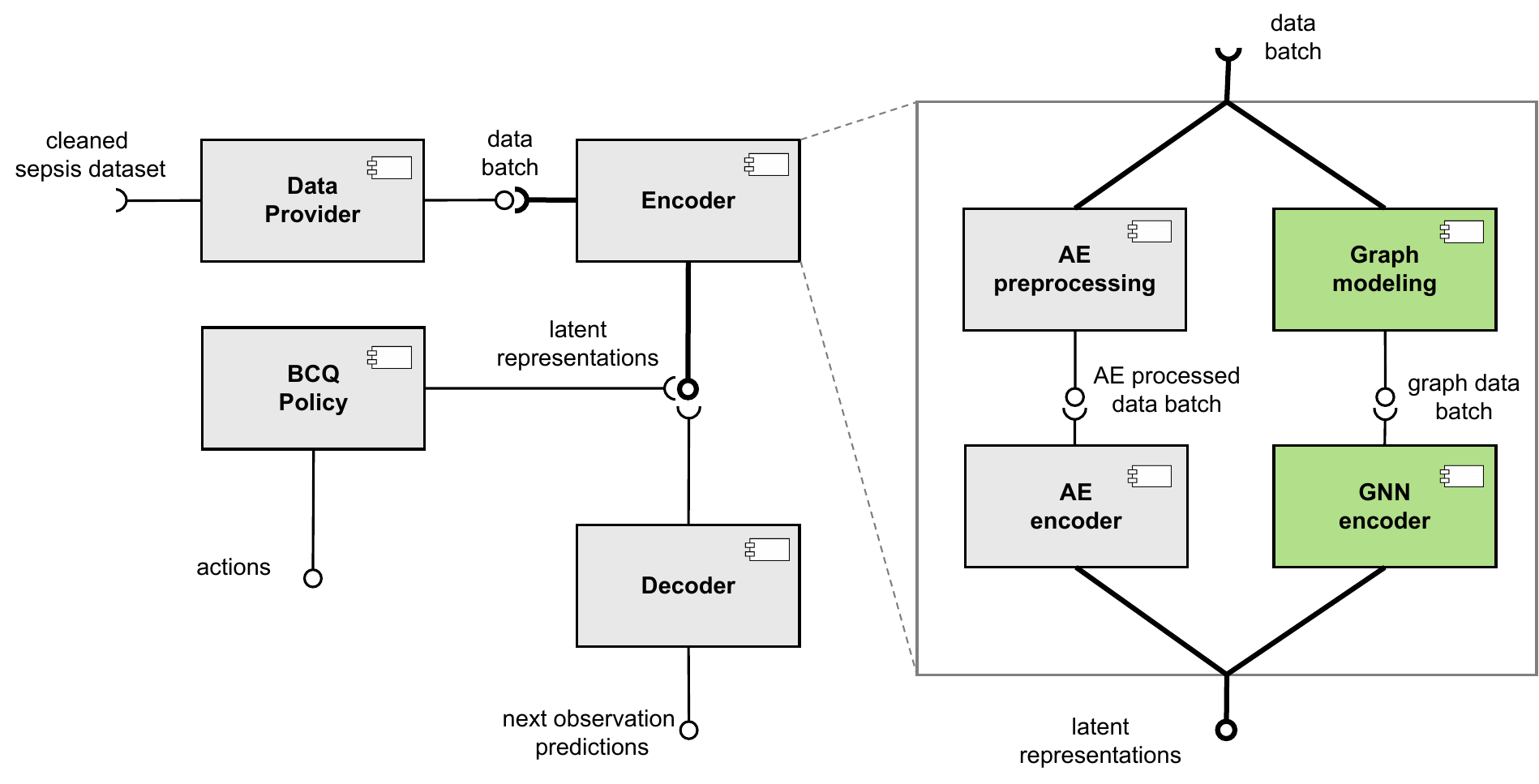}
	\caption{Component diagram of the proposed approach. The components representing our main contribution are shown in green.}
	\label{fig:componentdiagram}
\end{figure*}

\todo{Building blocks - Encoders}
The Encoder component is subdivided into two flows: the GNN flow and the AE flow (see \cref{fig:componentdiagram}). The GNN flow encompasses the Graph modeling component and the GNN encoder. These two components are the primary contribution of this paper and we describe them in the following subsections (see \cref{subsec:graph-modeling}, \cref{subsec:gnn-encoder}). The implementation of them the AE flow is reused from \cite{killian_empirical_2020}. The aim of the AE flow is to present a reference point to compare the GNN flow with. 

\todo{Building blocks - the rest components}
We also reuse the implementation of the Data provider, the Decoder and the BCQ policy to be able to compare with the results of \cite{killian_empirical_2020}. The BCQ policy is an abstraction for the BCQ algorithm. The role of the Data provider is to perform training, validation, and testing split with 70/15/15 proportion and then organize the data into batches of size $b$, with each batch containing $b$ trajectories.
After the batches are encoded into latent state representations, the Decoder  concatenates them with the next performed action and predicts the next observation of the patient. The BCQ policy inputs latent state representations as observations from the environment and outputs the treatment recommendations for clinitians.

\todo{Training phases}
Following Killian et al. \cite{killian_empirical_2020}, we decouple representation learning from policy learning. Thus, training contains two phases. In the first phase, we train autoencoders to predict features of the next patient observation, effectively training the AE encoder and the GNN encoders to extract meaning in the form of a latent vector. The autoencoders’ training process is detailed in \cref{subsec:gnn-pretraining}, and the results of the representation learning phase are presented in \cref{appendix:representation-learning}. In the second phase, the trained encoders are used to encode the data into latent representations. The representations serve as environment observations to train the BCQ policy. We evaluate the policy and present the results in \cref{sec:empiricalstudy}.  This setup enables a controlled, like-for-like comparison, meaning that any differences observed can be attributed to the representation (GNN vs. NN) rather that to changes in the BCQ. 

\subsection{Graph modeling}
\label{subsec:graph-modeling}

\todo{description of the section}
In this section, we explain the transformation process of tabular data into trajectory graphs. Obtaining trajectory graphs is an intermediate transformation. Each trajectory graph is processed further into sequences of graph snapshots to be used as an input to the GNN encoder (see \cref{fig:componentdiagram}). When modeling we take into account that exclusion of treatment history from the input data might lead to unsafe treatment recommendations \cite{lu_is_2020} and therefore we opted for inclusion of this information into each graph snapshot.

\todo{time-invariant and variant features}
In order to construct a trajectory graph, we divided the patient's features into two groups: time-invariant and time-variant features. We defined time-variant features as the features that change within one trajectory according to the data. We classified gender, age, readmission and ventilation statuses as time-invariant features. Unlike \cite{killian_empirical_2020} we consider a patient's body weight to be a time-variant feature, as it may change within one trajectory.

\begin{figure}[htbp]
	\centering
	\includegraphics[width=0.8\linewidth]{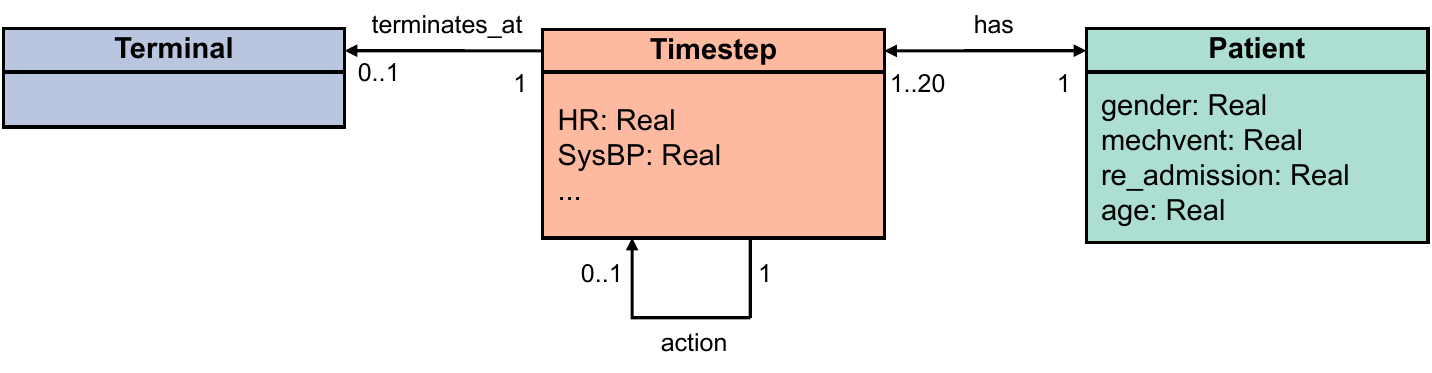}
	\caption{Type graph for patient trajectory graph.}
	\label{fig:typegraph}
\end{figure}

\begin{figure}[htbp]
	\centering
	\includegraphics[width=0.8\linewidth]{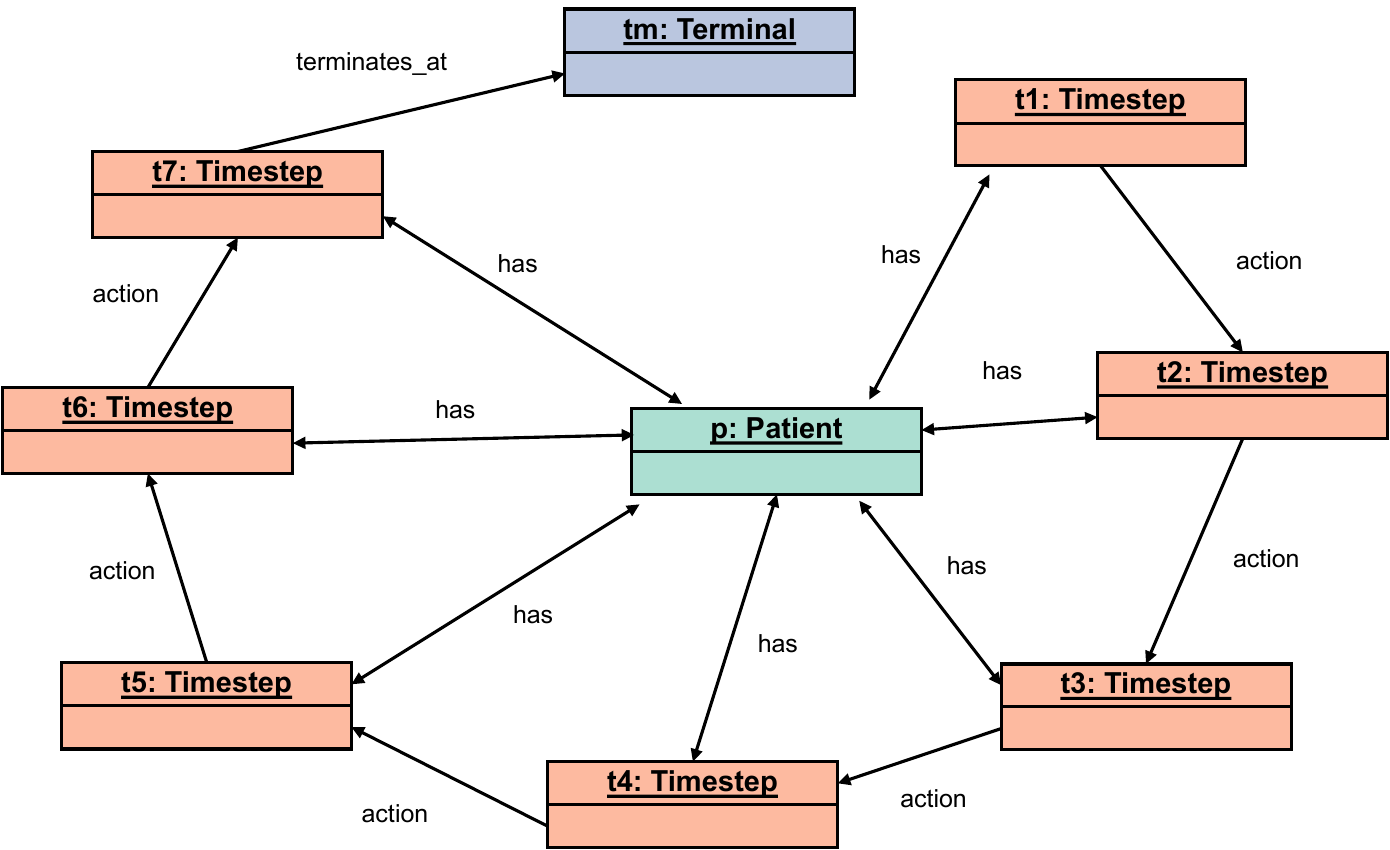}
	\caption{Patient trajectory graph with 7 time steps. }
	\label{fig:fulltrajectorygraph}
\end{figure}

\todo{graph modeling. trjaectory graph}
We modeled each patient trajectory as a dynamic heterogeneous graph. We describe the language of such dynamic heterogeneous graph using an attributed type graph, which is depicted in \cref{fig:typegraph}. An exemplary instance graph is shown in \cref{fig:fulltrajectorygraph}.  In  \cref{fig:typegraph} and \cref{fig:fulltrajectorygraph}, colors are used solely to facilitate the mapping of node types in instance graphs to the corresponding nodes types in the type graph. Time-invariant features $\mathcal{I}$ are encapsulated in a node type called Patient and time-variant features $\mathcal{T}$ – in Timestep nodes.  \cref{table:all-features} provides an overview of the features grouped into time-invariant and time-variant categories. Every patient trajectory has exactly one Patient node as a central element of the graph. Each Timestep node is connected to the next Timestep with a unidirectional edge that stores information about the action taken in-between the corresponding time steps. \textcolor{black}{Each such edge has a feature vector represented as a one-hot encoding of length 25, containing all zeros except for a single one indicating the action taken. The Patient node is connected to each Timestep node via bidirectional edges, with all edge weights set to 1. }The aim is to represent the Patient-Timestep association and to avoid introducing any other domain-specific information by using unequal weights. The bidirectional edges allow messages to be passed into both directions later at the GNN encoding stage. Terminal-type nodes indicate the end of the trajectory and are therefore connected with the last Timestep node in each trajectory. A Terminal node contains information on a reward for an RL agent: 1 if the patient survived and -1 if not (see \cref{subsec:sepsis-mdp}). \textcolor{black}{We introduced Terminal nodes for convenience of the subsequent transformation; the edge connecting the final Timestep to the Terminal node carries no attributes.}

\todo{Graph snapshots}
The subsequent stage of the transformation process involves  graph snapshots generation from each trajectory graph (see \cref{fig:graphsnapshots}). Each snapshot $g_t$ contains information about the patient's current observation, as well as their previous observations and treatment history. A snapshot $g_t$ comprises a Patient node from the trajectory graph and Timestep nodes from the initial time point up to the current time point. Each $g_t$ is encoded at the next stage to a latent representation $l_t$. A batch of graphs refers to a certain number of sequences of graph snapshots.

\subsection{GNN encoder}
\label{subsec:gnn-encoder}

The GNN encoder component produces a GNN designed to work with heterogeneous graphs containing different types of nodes and edges. The encoder consists of several heterogeneous graph convolution layers, followed by pooling operations and a linear layer. The final output is a fix-size latent representation for each graph. We use 2 alternative convolution layer types: SAGEConv operators from the GraphSAGE framework (henceforth SAGE) \citep{sageconv_hamilton_inductive_2017} and the GATv2Conv operator from the GATv2 architecture\citep{gatv2_brody_how_2022}. 
The variable parameters for the GNN are the number of output features $f_{out}$ in convolutional layers and the number of convolutional layers $n_{conv}$.

\paragraph*{SAGEConv}
During the forward pass, the input passes through the $n_{conv}$ convolution SAGEConv layers, where each layer contains 3 SAGEConv operators corresponding to the number of edge types. Each layer is followed by a ReLU \cite{relu} activation function. The features are aggregated using 'mean' for each edge type, which means that if a node has more than one incoming edge of the same type, then these incoming messages are averaged. In our case, this parameter is relevant for the Patient node, because this node has many incoming edges of the same type from Timestep nodes, so, messages from each Timestep are averaged. After that, features from different edge types are summed up to accumulate information from different types of relationships. This step is important for Timestep nodes, as Timestep nodes have 2 different types of incoming edges.

After the convolution layers, graph-level mean pooling is applied in order to aggregate node features for each node type. As a result, we get a feature vector per node type. As we have two node types, Timestep and Patient, after the graph-level mean pooling we obtain two feature vectors. 
The next step is to aggregate them with the summation over different node types in order to get one vector. The size of this vector is defined by the hyperparameter $f_{out}$. As a next step, we add a linear layer of size $(f_{out}, l)$ to ensure that the output vector is of length $l$ even if $f_{out} \neq l$.

\paragraph*{GATv2Conv}
The second architecture uses the GATv2Conv operator in the layers. The difference with the SAGEConv architecture is that this operator (1) takes into account the weight of the edges in the input graph, (2) has additional training weights that indicate the importance of each neighbour node. This means that instead of using a simple average aggregation for each edge type, GATv2Conv calculates the weighted average. 

%% file: experiments.tex
\section{Empirical Study}

\label{sec:empiricalstudy}
\todo{Research Question}
Our aim is to answer the question of how using representations derived from graph-structured data via trained GNN encoders influences the accuracy and efficiency of policy learning. To answer this question, we train GNN and AE encoders, along with a decoder, to predict the next patient's state. The downstream task for the encoders, as derived from the research question, is policy learning by a RL agent using the dBCQ. We evaluate the learned policies using WIS.

\todo{Common info}
 The hyperparameters that are common to both the autoencoders training of the encoders and to the RL policy learning are $l$, the size of the latent state representation vectors, which is set to 64; and $b$, the batch size, which is set to 128.

\subsection{AE and GNN autoencoders training and hyperparameters}
\label{subsec:gnn-pretraining}

The objective of an autoencoder is to reconstruct the vector of time-variant patient features $c_{t+1}$, a subset of the observation $o_{t+1}$ available in the data (see \cref{subsec:graph-modeling}).The reconstruction is based on the current observation $o_t$, the preceding action $a_{t-1}$ (or their history in the case of GNNs), and, at the decoder stage, the subsequent action $a_t$.

The goal of training is to obtain encoders that extract meaningful information from a patient observation in the form of a latent vector of size $l$. The training process work as follows: each encoder receives the current observation $o_t$ togther with the preceding action $a_{t-1}$. Based on this input, the encoder produces a latent representation vector of size $l$, which is concatenated with the subsequent action $a_t$ known from the data. The resulting vector serves as input to the decoder, which outputs a prediction of time-variant features $\hat{c}_{t+1}$ of the next patient observation. The loss is calculated as the differnce between the true time-variant patient features  $c_{t+1}$ and the predicted features $\hat{c}_{t+1}$.

\todo{Decoder}
\paragraph*{Decoder}
While the encoder architectures differ, the decoder architecture remains nearly identical across all considered encoders. The decoder consists of three fully-connected layers $(l+25, 64), (64, 128), (128, \mathrm{obs\_dim}) $ . For the GNNs case, $\mathrm{obs\_dim=34}$, and for the AE case, $\mathrm{obs\_dim=33}$. The hidden layers are followed by a ReLU activation function. The final layer of size obs\_dim produces a vector that serves as the mean of a unit-variance multivariate Gaussian distribution. This is then used to predict the time-variant features of the next patient observation $\hat{c}_{t+1}$.

\todo{AE encoder}
\paragraph*{AE encoder}
For the AE encoder, we use the optimal hyperparameters from previous work \cite{killian_empirical_2020}. It is a three-layered, fully connected NN with layer sizes of: $(\mathrm{obs\_dim}+25, 64), (64, 128)$ and $(128, l)$. The ReLU activation function is applied to the hidden layers. We train it for 600 epochs using the Adam optimizer with a learning rate of $5\mathrm{e}{-4}$.

\todo{GNN encoders}
\paragraph*{GNN encoders}
We conduct an empirical study to determine the optimal hyperparameters for the GNN encoders. As for tuning the hyperparameters of the SAGEConv encoder, we set $(f_{out}$,  $n_{conv}) \in \{2, 3\} \times\{64,128\}$, i.e., $(f_{out}, n_{conv}) = (2,64),\ (2,128),\ (3,64),\ (3,128)$. The best-performing model is found at $(f_{out}^{*}, n_{conv}^{*}) = (2,64)$. Setting $f_{out}$ to 128 results in clear overfitting. Therefore, for GATv2Conv, we fix  $f_{out}^{*}$ to 64 and vary $n_{conv}$ within the range $ \{1, 2, 3\}$. The best performing model has  $n_{conv}^{*} = 1$. We train the GNN autoencoders for 200 training epohs using the Adam optimizer with a learning rate of $1\mathrm{e}{-3}$. \cref{appendix:representation-learning} contains the visualized training results. An overview of the selected hyperparameters is provided in \cref{table:gnns-final-hyperparameters}.

Although all autoencoders share same training principle, they differ fundamentally in how they handle input data. The AE encoder has no access to the history of past observations or treatments: it processes each time step independently by concatenating the previous action $a_{t-1}$ with the current observation $o_t$. Any information about observations at a time step $\le t-1$ are not available to the network (see \cref{subsec:threats}). By contrast, GNN encoders operate on snapshot graphs $g_t$ (see \cref{subsec:graph-modeling}). Each snapshot is still processed independently, but its graph structure inherently contains all observations and actions from time $1$ up to $t$. This allows GNNs to incorporate historical information without explicit recurrence.

\subsection{dBCQ-Policy Learning and Evaluation}
\label{subsec:policy-learning}

We trained policies using data encoded by the three best performing trained encoders: AE, GNN-SAGE and GNN-GATv2. The policy models were trained using the dBCQ algorithm with identical hyperparameters, and the evaluation is performed using WIS (see \cref{sec:prelim}). 

\todo{dBCQ}

The dBCQ model was trained on the combined training and validation subsets of the data for one million iterations. The WIS evaluation was performed every 500 iterations on the testing subset. All subsets (training, validation, and testing) were first encoded into latent representations using the selected encoder. The Q-network consisted of three fully connected layers, with 64 nodes in each of the two hidden layers, followed by ReLU activations, and 25 nodes in the output layer. The BCQ action elimination threshold was set to 0.3, as in related work \cite{killian_empirical_2020}. The learning rate was set to \(1 \times 10^{-3}\) for all experiments. An overview of the selected hyperparameters is provided in \cref{table:bcq-final-hyperparameters}.

\todo{Behav cloning}
To perform the WIS evaluation, we first trained clinician-like behavioral policies documented in the data. We used a neural network with three fully connected layers \((38, 128), (128, 128), (128, 25)\), trained with cross-entropy loss. ReLU activations were applied to the hidden layers, and batch normalization was employed to regularize training. The supervised task consisted of imitating clinicians’ actions based on patient observations. Since the actions were discretized (see \cref{subsec:sepsis-mdp}), this was formulated as a classification problem. Training was carried out with the Adam optimizer at a learning rate of \(1 \times 10^{-4}\). The behavioral policy was trained directly on the raw data, without encoding into latent representations. The set of hyperparameters used in our experiments is summarized in \cref{table:bc-final-hyperparameters}.

\begin{figure}[htb]
	\centering
	\includegraphics[width=0.8\textwidth]{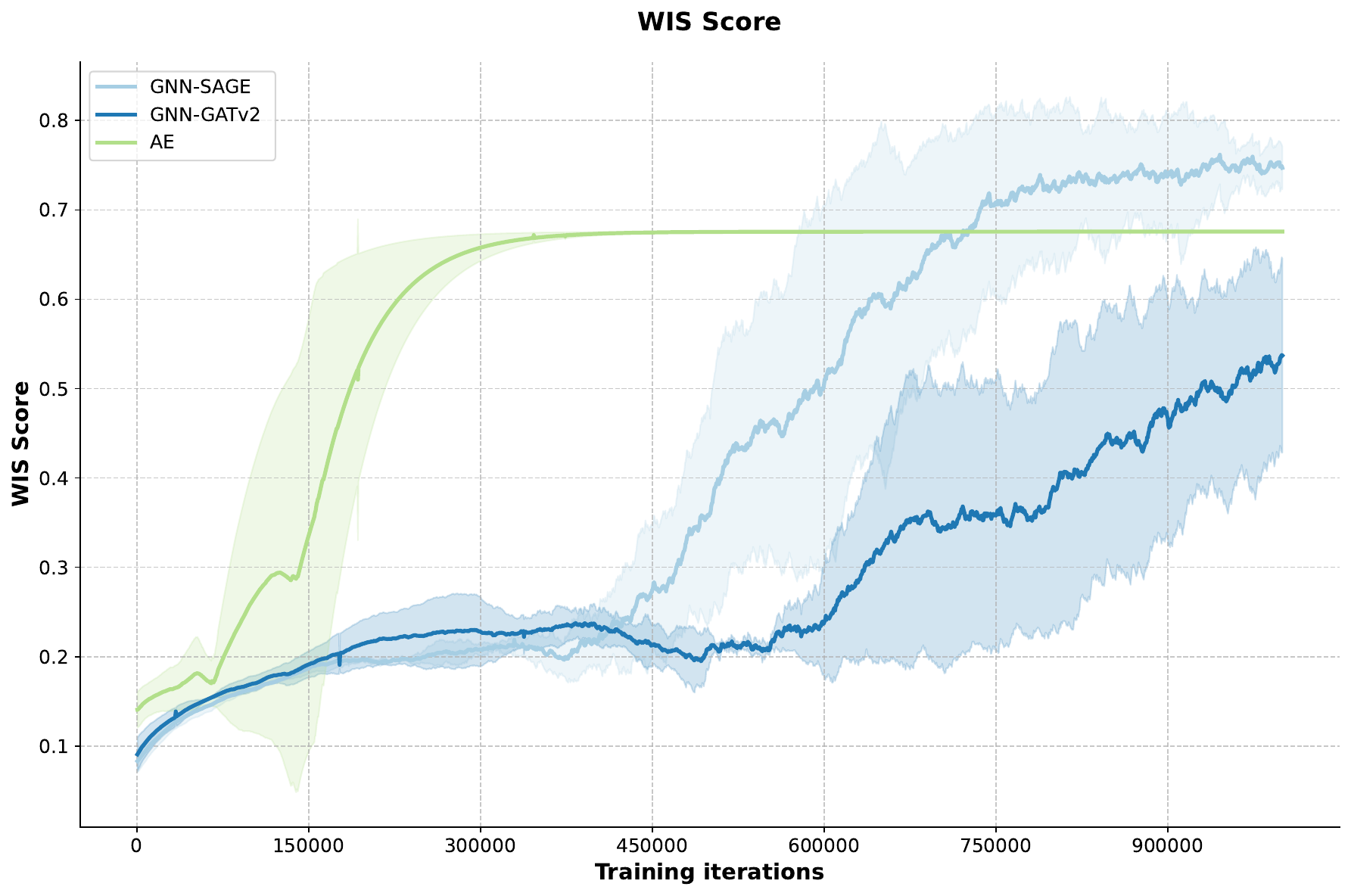}
	
	\caption{WIS on 1e6 training iteration on GNN-SAGEConv and GNN-GATv2Conv}.
	\label{fig:wis-normal-1e6}
	
\end{figure}

\todo{Results of WIS}
We run the experiments three times for each of the policy experiment curves, using the random seeds 1234, 2020, and 2025. We apply an exponential moving average with smoothing parameter $alpha=0.1$. We then calculate the mean and standard deviation, plotting the results. \cref{fig:wis-normal-1e6} demonstrates the results of the WIS evalution of the policies learnt from observations obtained from different trained encoders – AE, GNN-SAGE and GNN-GATv2. We set the number of training iterations to 1M. Initially, we trained on 500K (see \cref{fig:wis-normal-5e5}), but this number of iteration was not sufficient for the GNN cases to show any significant growth. The GNN-SAGE curve shows steady growth from 375K iterations, while GNN-GATv2 - starts to show growth only after 600K iterations. The AE learning curve demonstrates significatly faster growth than both GNN representation types: it demonstrates a rapid increase at 75K iterations. However, the WIS score for AE representations plateaux at 0.68 after around 350K iterations, while the score for GNN-SAGE representation grows to 0.75 by the end of the training. GNN-GATv2 representations show the worst performance with respect to both accuracy and efficiency, as steady growth only begins after 600K iterations, and by the end of the training, the WIS score is just above 0.5.

\subsection{Discussion}
\label{subsec:discussion}

\todo{Compare with Killian et al.}
Compared to Killian et al. \cite{killian_empirical_2020},  we successfully reproduced the AE learning curve, although their plot covers 200K iterations while ours extends to 1M.  As Killian et al. do not present results beyond 200K, it remains unclear how performance evolves, and our work provides an inside into what happens to the AE curve on the later iterations. However, their work clearly demonstrates that policies learned from recurrent encodings – especially CDE – converge much faster than those based on GNNs. In their results,  the CDE curve plateaux at a WIS score of approximately 0.78 after 75K iterations. In contrast, the GNN-SAGE curve requires significantly more steps to begin showing steady improvement, but it ultimately demonstrates competitive performance, reaching a WIS score of approximately 0.75. This suggests that, when sufficiently trained, a GNN-SAGE based policy can match – or nearly match – the acccucy of a CDE based policy. Therefore , our future work will investigate encoders based on a combination of RNNs and GNNs.

\todo{Accuracy on pretraining vs accuracy on downstrem task}
The accuracy results for autoencoders training did not correspond to those for the downstream policy learning task. AE and GNN-SAGE performed worse than GNN-GATv2 in the autoencoders training task, but better in policy learning. While the difference in autoencoders training results between AE and GNN-SAGE was marginal, the shapes of their respective policy training learning curves were very different.

\todo{Complex models - worse performace }
It appears that the additional learned weights in GNN-GATv2, which represent the importance of each node neighbour, might mislead the policy learning agent, thereby slowing down the learning curve compared to GNN-SAGE, where edge weights are neither taken into account nor learned. This is consistent with the findings of Killian et al., who discovered that more complex models do not necessarily result in a better learning curve.

\todo{Hardware info. Issue with efficiency }
The experiments were conducted on a system comprising an AMD EPYC 9554 processor with 128 cores running at 3.76 GHz, three NVIDIA L40 GPUs each with 46 GB of memory, and 1.5 TB of system RAM. The system runs CUDA version 12.7. Training the BCQ policy on the GPU took approximately three hours for each experiment with 500K iterations, and around six hours with 1M iterations. Traning the autoencoders on the GPU for each GNN took around 30 hours, whereas AE took between three and four hours, around ten times faster. We trained the models using both the CPU and the GPU. For GNN autoencoders training, using the GPU does not speed things up. One possible reason for this is that the graphs we use are relatively small; this issue will be addressed in future work.

\subsection{Threats to Validity}
\label{subsec:threats}

This section discusses threats to validity.  Specifically, we consider 4 aspects of validity–construct, internal, external, reliability \cite{threats_to_validity_wohlin_experimentation_2024}–which we outline here and aim to address in future work.

\paragraph*{Construct Validity}
We identify several construct‐validity threats. First, because our goal was to study GNNs, we converted the data into a graph, but this was only one possible modeling approach, so it remains unclear how GNNs would perform with a different data representation. Second, we evaluated learned policies using a quantitative metric–WIS score, which, on the one hand, was used in related work to evaluate sepsis treatment policies leaned in the offline RL settings \cite{killian_empirical_2020}, but, on the other hand, we did not perform any qualitative assessment, so it is uncertain whether high scores truly reflect clinically meaningful decisions. Finally, to estimate the RL policy’s value via WIS, we must first learn a NN approximation of the clinitians’ behavior policy; because this surrogate is imperfect, the importance weights are biased and the estimated policy values may not reflect true performance of the RL policies.

\paragraph*{Internal Validity}
We address an internal‐validity threat by decoupling representation learning from policy learning. This isolation ensures that any performance differences can be attributed to the encoder itself rather than to entangled training dynamics. Another internal-validity concern arises from our experimental setup: the GNN models utilize historical data directly incorporated into the graph structure, whereas the AE does not. This difference makes it challenging to isolate the architectural effects from simply having more input information. Nonetheless, the results show, that even when both SAGE and GATv2 encoder architectures received identical inputs (including history), their performances diverged–SAGEConv outperformed AE, while GATv2 underperformed relative to AE. These mixed results suggest that architectural differences may matter more than mere access to historical data. Finally, to reduce random error in the WIS analysis, we ran each experiment multiple times, averaged the results, and reported the standard deviations.

\paragraph*{External Validity}
We addressed external‐validity threats by using the standard evaluation metric–WIS, which is common in RL‐based sepsis treatment and allows comparison with the previous study of Killian et al. \cite{killian_empirical_2020}. However, since we only used the MIMIC‐III dataset, it remains unclear whether our results would generalize to other datasets—this limitation necessitates further investigation.
	
\paragraph*{Reliability}
In terms of reliability, we consistently used the same stratified training, validation, and test splits throughout our experiments; however, because we did not fix a random seed when creating those splits, a future researcher attempting to reproduce our work could obtain different partitions. We mitigated imbalance by stratifying on trajectory outcomes, ensuring each subset contained proportional positive and negative cases. Regarding random NN initialisation for the RL policy learning, we set and report random seeds. To further reduce variability, we report mean results of multiple runs and standard deviations, providing transparency about stability. Finally, \cref{fig:wis-normal-1e6} shows that the results of policy training from the AE representations match those of Killian et al. \cite{killian_empirical_2020} for the first 200K training iterations, despite a different data split, supporting the reliability of our findings.

%% file: relatedwork.tex
\section{Related work}
\label{sec:relatedwork}

\todo{GNNs in healthcare in general}
GNNs are used in the context of representation learning in the heathcare domain \cite{wanyan_deep_2021, rocheteau_predicting_2021}. The survey  \citep{graph_healthcare_survey} shows that there has been an increase in the number of studies using GNN for clinical risk prediction since 2020. Unlike us, most of the papers focus on diagnosis prediction rather than treatment policy. The most popular architecture used is GAT \citep{gat_velickovic_graph_2018}.

\todo{Sepsis treatment using RL}

One of the first mature studies on the RL applied to sepsis treatment was published by Komorowski et al. \cite{komorowski_artificial_2018}. They used a discrete action space for treatment recommendations, which was adopted by many subsequent studies, as summarised by Roggeveen et al. \cite{roggeveen_transatlantic_2021}.
Huang et al., on the other hand utilize continuous action spaces \cite{contin-act-space-sol}. Komorowski et al.\cite{komorowski_artificial_2018} used a discrete state space by applyng clustering to patient states and utilizing tabular Q-learning. However, most later works use continuous state spaces and apply deep RL \cite{dddqn_2024, zhang_optimizing_2024, wu_value-based_2023, tamboli_reinforced_2024, liang_treatment_2023}. Killian et al. and Huang et al. encoded patient states into latent state observations \cite{killian_empirical_2020, contin-act-space-sol}. The reward in related work is either terminal, based on the survival of the patient, or includes intermediate signals based on short-term changes in the patient's condition, with no clear preference for either variant. The field of RL for sepsis treatment is continually being explored, with new studies appearing every year \cite{tu_offline_2025}.

\todo{GNN and RL}
GNNs have also been successfully combined with RL in domains such as Google Research Football \cite{football_niu_graph_2022} and routing optimization \cite{almasan_deep_2022}. In contrast to our work, these studies optimized the RL policy and the representation jointly through end-to-end training.

\todo{GNN in sepsis domain}
To our knowledge, there are no papers that use GNNs to find an optimal sepsis treatment strategy. In the domain of sepsis, there was a study that focuses on predicting the onset of sepsis \cite{10020963}. They also model medical data as a graph and use GNNs to learn the representations. The dataset they use is not openly available.

%% file: conclusion.tex
\section{Conclusion and Future Work}
\label{sec:conclusion}

We applied a graph‐based approach to optimize sepsis treatment with reinforcement learning. Patient data from MIMIC‐III were modeled as dynamic heterogeneous graphs, and we trained GNN based autoencoders (based on the GraphSAGE and GAT architectures) to generate latent state representations. Following prior work, representation learning was decoupled from policy learning and the learned representations were used to train an offline RL agent. Our results show that the GraphSAGE encoder outperforms traditional relational methods in accuracy and is comparable to the (recurrent‐encoding) prior approach, although it requires more training iterations to converge.

In future work we will address further threats to validity from~\cref{subsec:threats}. Firstly, we will perform a qualitative analysis of the learned policy for comparison with other approaches, and will evaluate its clinical interpretability. Secondly, we will explore alternative graph modeling approaches. Specifically, we will distribute the patient features to multiple nodes instead of modeling them as attributes of one node. This should address the issue of long training time. Finally, we plan to explore dynamic GNN encoders that incorporate recurrent neural networks in order to better capture temporal dependencies.

%% file: appendix.tex
\begin{appendices}

	\section{Representation Learning Phase}
	\label{appendix:representation-learning}
	 \textcolor{black}{\crefrange{fig:ae-seeds-loss}{fig:inject_action_experiment}  were produced with the visualization tools provided by Weights \& Biases \cite{wandb}}. The name of each run in the figures is automatically generated and provides no additional information other than being a unique, human-readable run identifier.

	Because obs\_dim varies depending on the encoder type (see \cref{subsec:gnn-pretraining}), we normalize the loss when comparing the encoders performance . For both training and validation sets, we compute the loss per batch and average over all batches. Finally, each average batch loss is divided by the batch size to obtain an average trajectory loss.

	\subsection{AE Encoder Training}
	
\begin{figure}[htb]
		\centering
		\begin{subfigure}[b]{0.45\textwidth}
			\includegraphics[width=\textwidth]{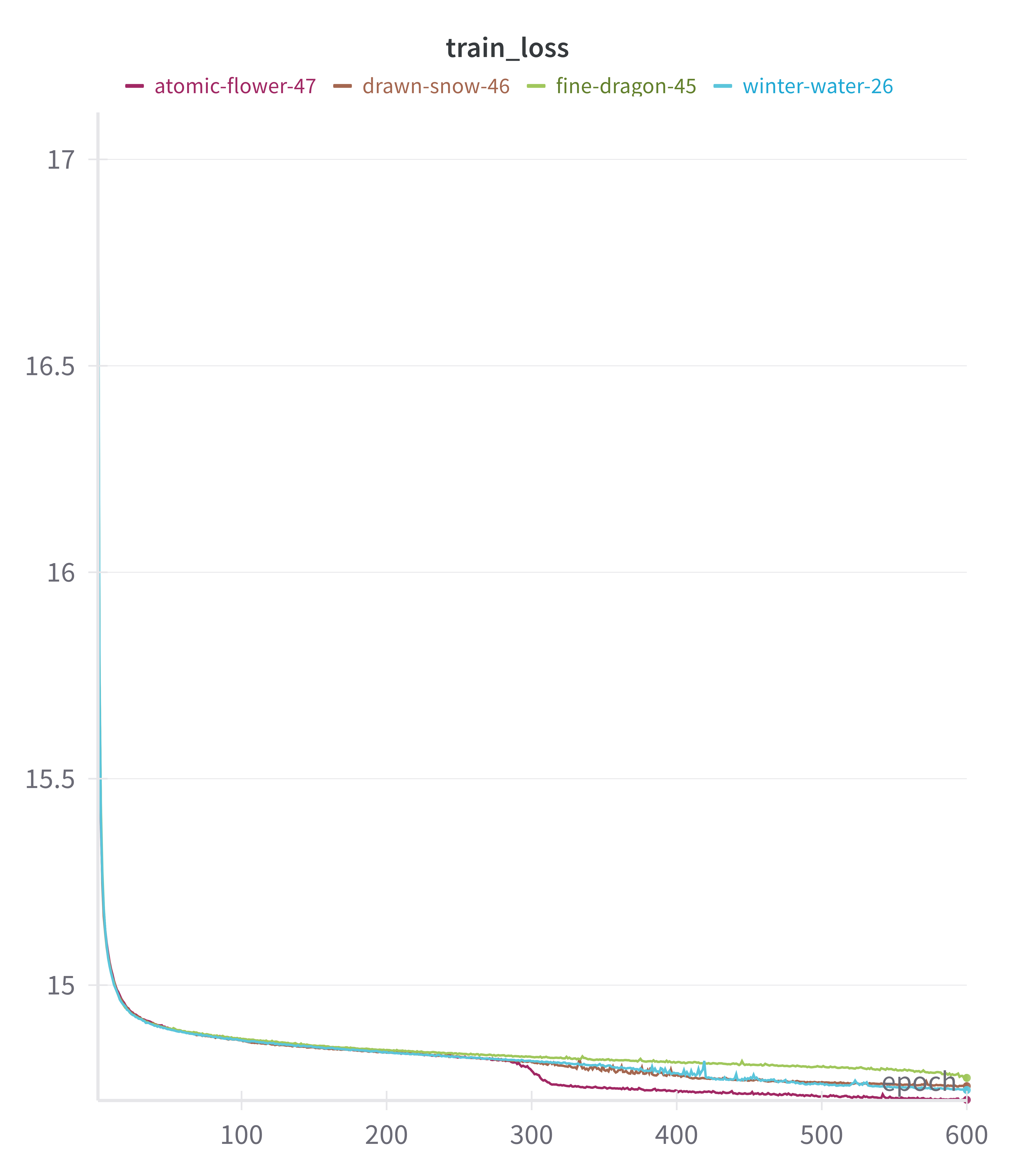}
			\caption{Training loss.}
			\label{fig:ae-seeds-train-loss}
		\end{subfigure}
		\hfill
		\begin{subfigure}[b]{0.45\textwidth}
			\includegraphics[width=\textwidth]{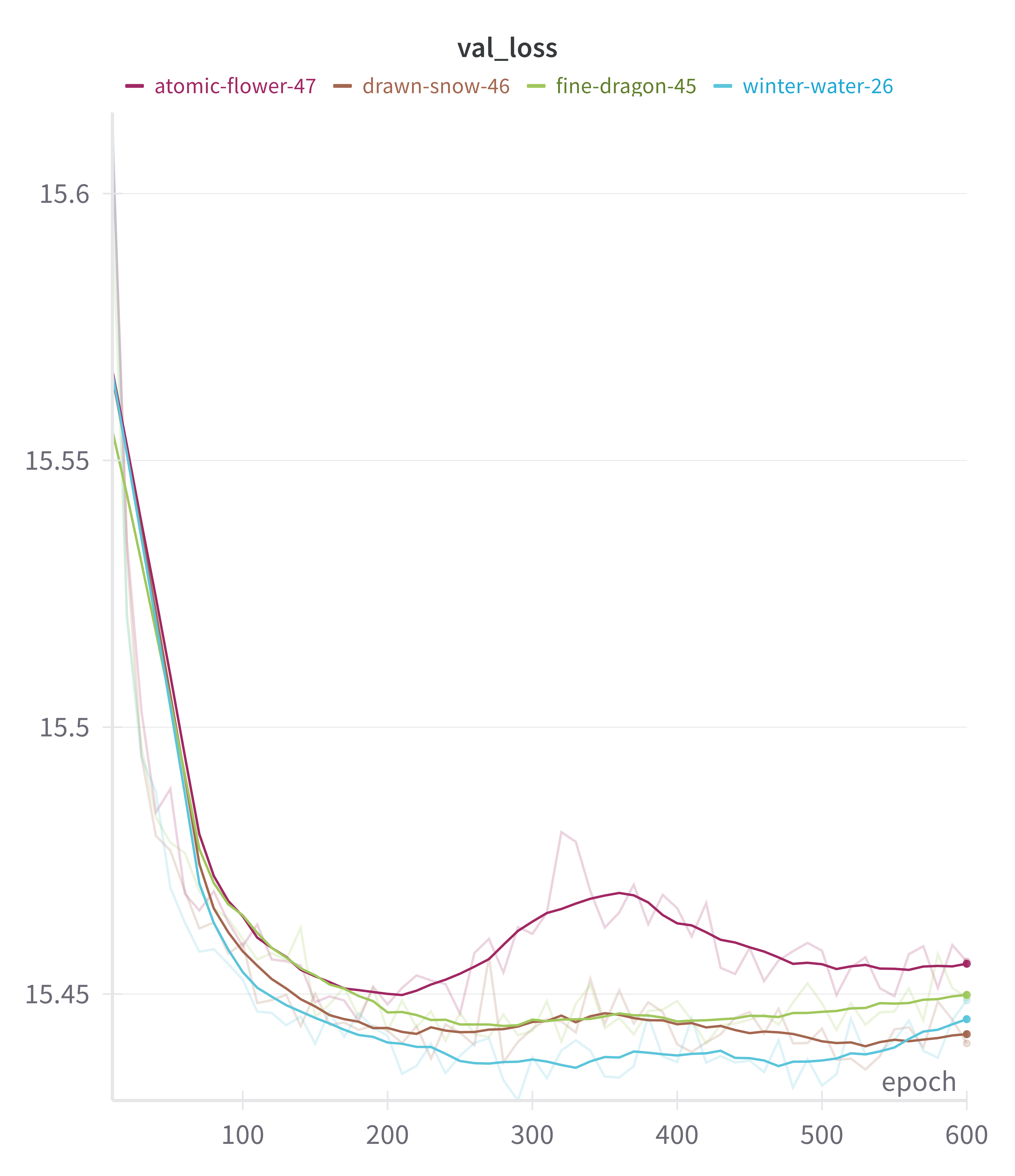}
			\caption{Validation loss.}
			\label{fig:ae-seeds-val-loss}
		\end{subfigure}
	
		\caption{Training AE autoencoder with four different seeds.}
		\label{fig:ae-seeds-loss}
	\end{figure}

	To replicate Killian et al.’s AE training \cite{killian_empirical_2020}, we adopted their hyperparameters: 600 training epochs, a learning rate of $5\times10^{-4}$, and validation every 10 epochs. The AE was run with four random seeds (1234, 25, 32, and 2020); the resulting loss curves are shown in \cref{fig:ae-seeds-loss}.
	
	The loss reported (see \cref{fig:ae-seeds-loss}) is the average trajectory loss. No smoothing was applied to the training loss; however, the validation loss was smoothed using a 10-step running average. Of the four runs, one began to overfit around epoch 30, but its validation loss resumed declining after peaking at approximately epoch 350. By epoch 600, the average validation loss across all runs was 15.448765.
	
	\subsection{GNN Encoder Training}
	
	Both GNN‐SAGEConv and GNN‐GATv2Conv were trained for 200 epochs. For visualization, we smoothed the validation curves using a 10‐step running average. Only encoder hyperparameters varied between runs; decoder hyperparameters remained fixed. \cref{table:gnns-final-hyperparameters} provides the overview of the GNN encoders hyperparameters.
	
	\begin{table}
		\caption{Selected hyperparameters for the GNN encoders.}
		\label{table:gnns-final-hyperparameters}
		\centering
		\begin{tabular}{lllllll}
			\toprule
			Encoder & Epochs& lr & $f_{out}^{*}$ & $n_{conv}^{*}$ & Intra aggr.& Inter aggr. \\
			\midrule
			GNN-SAGEConv   & 200 & $1\mathrm{e}{-3}$ & 64 & 2 & mean & sum         \\
			GNN-SAGEConv   & 200 & $1\mathrm{e}{-3}$ & 64 & 1 & attn. & sum         \\
			
			\bottomrule
		\end{tabular}
	\end{table}
	
	\paragraph*{GNN-SAGEConv}
	
\begin{figure}[htb]
		\centering
		\begin{subfigure}[b]{0.45\textwidth}
			\includegraphics[width=\textwidth]{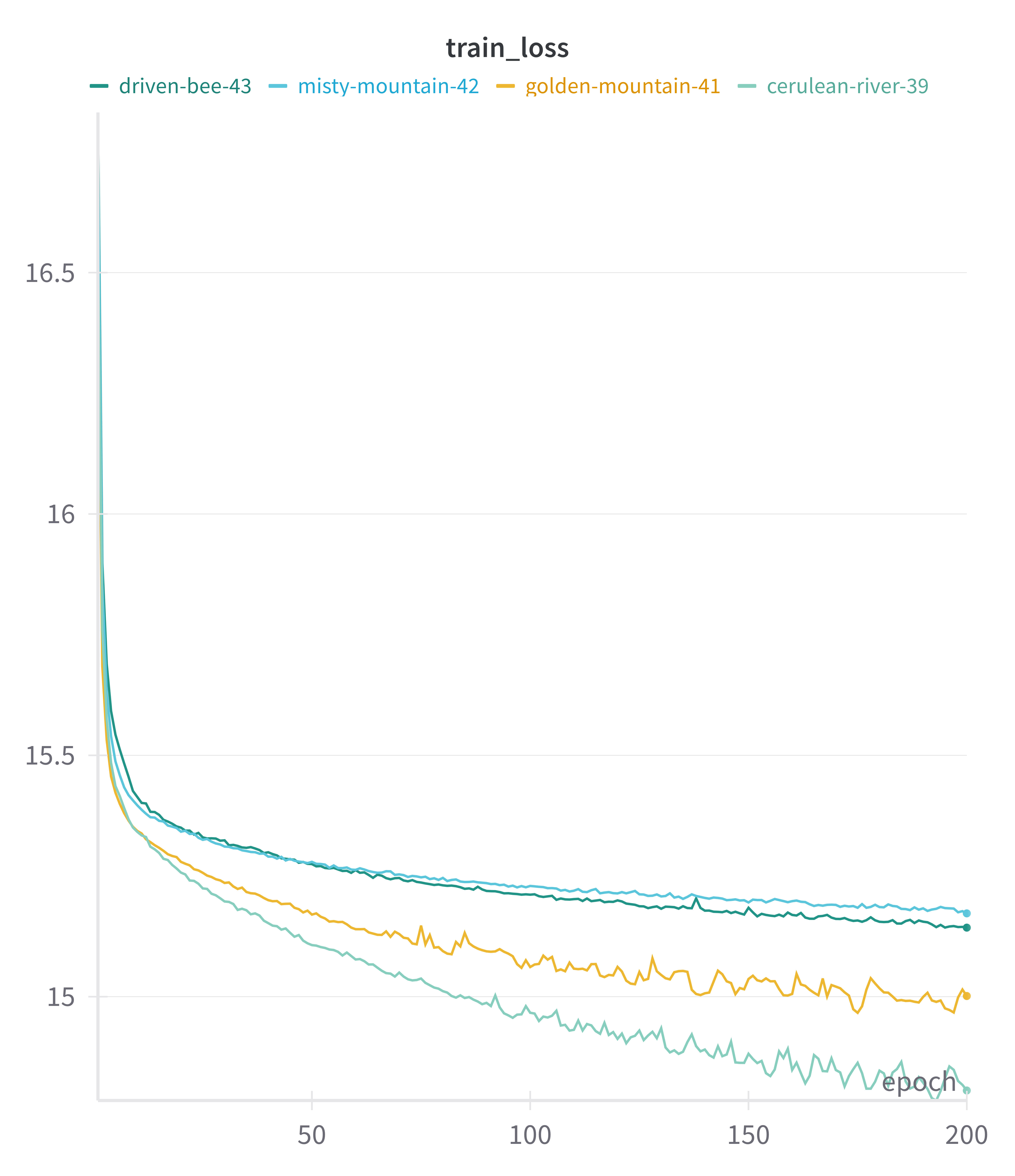}
			\caption{Training loss.}
			\label{fig:gnn-sage-train}
		\end{subfigure}
		\hfill
		\begin{subfigure}[b]{0.45\textwidth}
			\includegraphics[width=\textwidth]{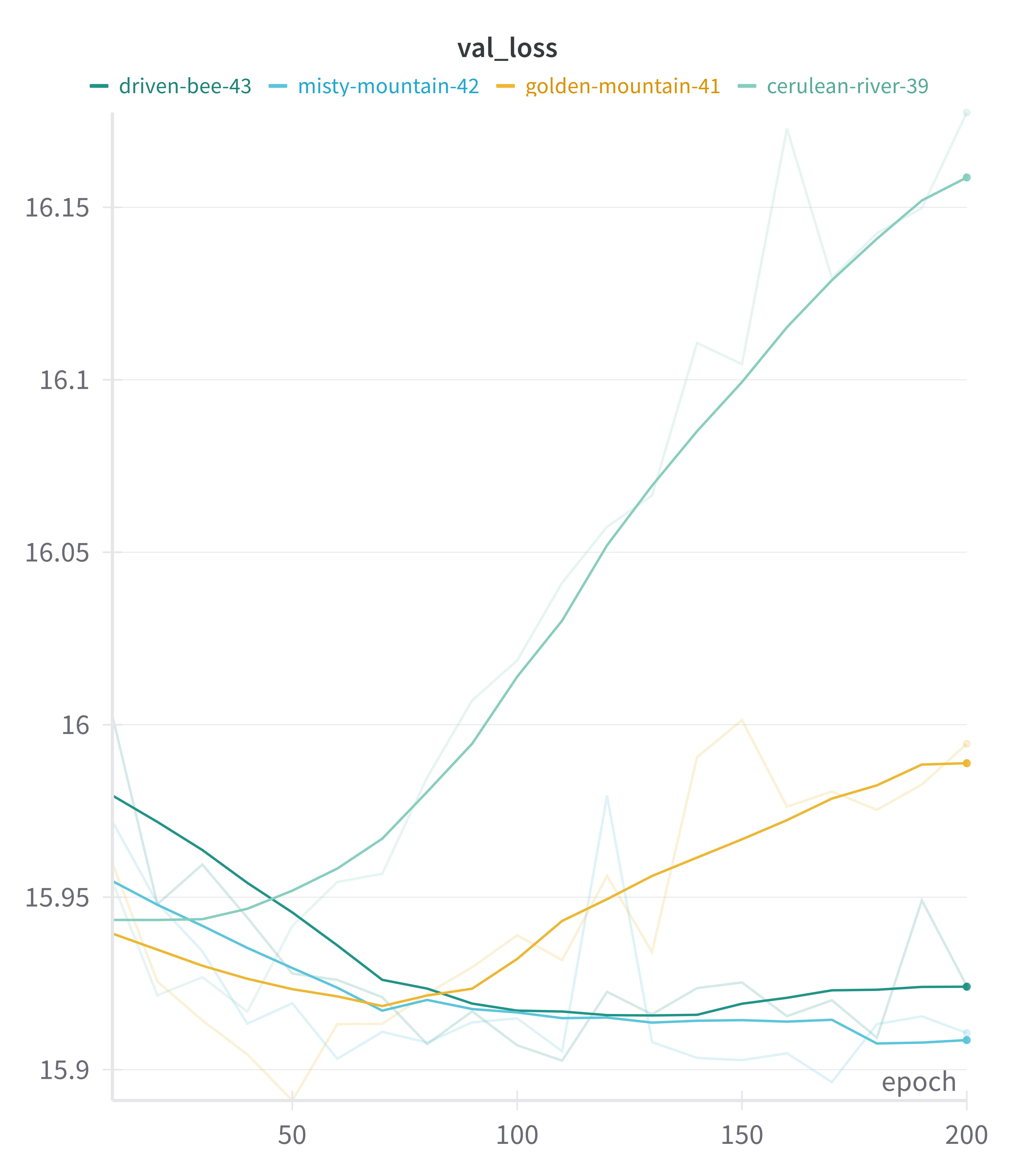}
			\caption{Validation loss.}
			\label{fig:gnn-sage-val}
		\end{subfigure}
	
		\caption{Training GNN-SAGEConv encoder.}
		\label{fig:gnn-sage}
	\end{figure}

	
	GNN‐SAGEConv was trained under four hyperparameter combinations:  $f_{out} = \{64, 128\}$ and $n_{conv} = \{2, 3\}$ (\cref{table:gnn-sage}). The best model—lowest validation loss—used $(f_{out}, n_{conv}) = (64, 2)$ (\cref{fig:gnn-sage}). Increasing the number of features in the graph convolutional layers to 128 led to overfitting for the \textit{cerulean-river-39} and \textit{golden-mountain-41} runs. Choosing the lower number of features and higher number of layers slightly increased the validation loss.

	\begin{table}
		\caption{Considered hyperparameters combinations for the GNN‐SAGEConv.}
		\label{table:gnn-sage}
		\centering
		\begin{tabular}{lll}
			\toprule
			Run name             & $f_{out}$ & $n_{conv}$ \\
			\midrule
			golden-mountain-41   & 128       & 2          \\
			cerulean-river-39    & 128       & 3          \\
			driven-bee-43        & 64        & 3          \\
			\textbf{misty-mountain-42} & \textbf{64} & \textbf{2} \\
			\bottomrule
		\end{tabular}
	\end{table}

	
	\paragraph*{GATv2Conv}
	
\begin{figure}[htb]
		\centering
		\begin{subfigure}[b]{0.45\textwidth}
			\includegraphics[width=\textwidth]{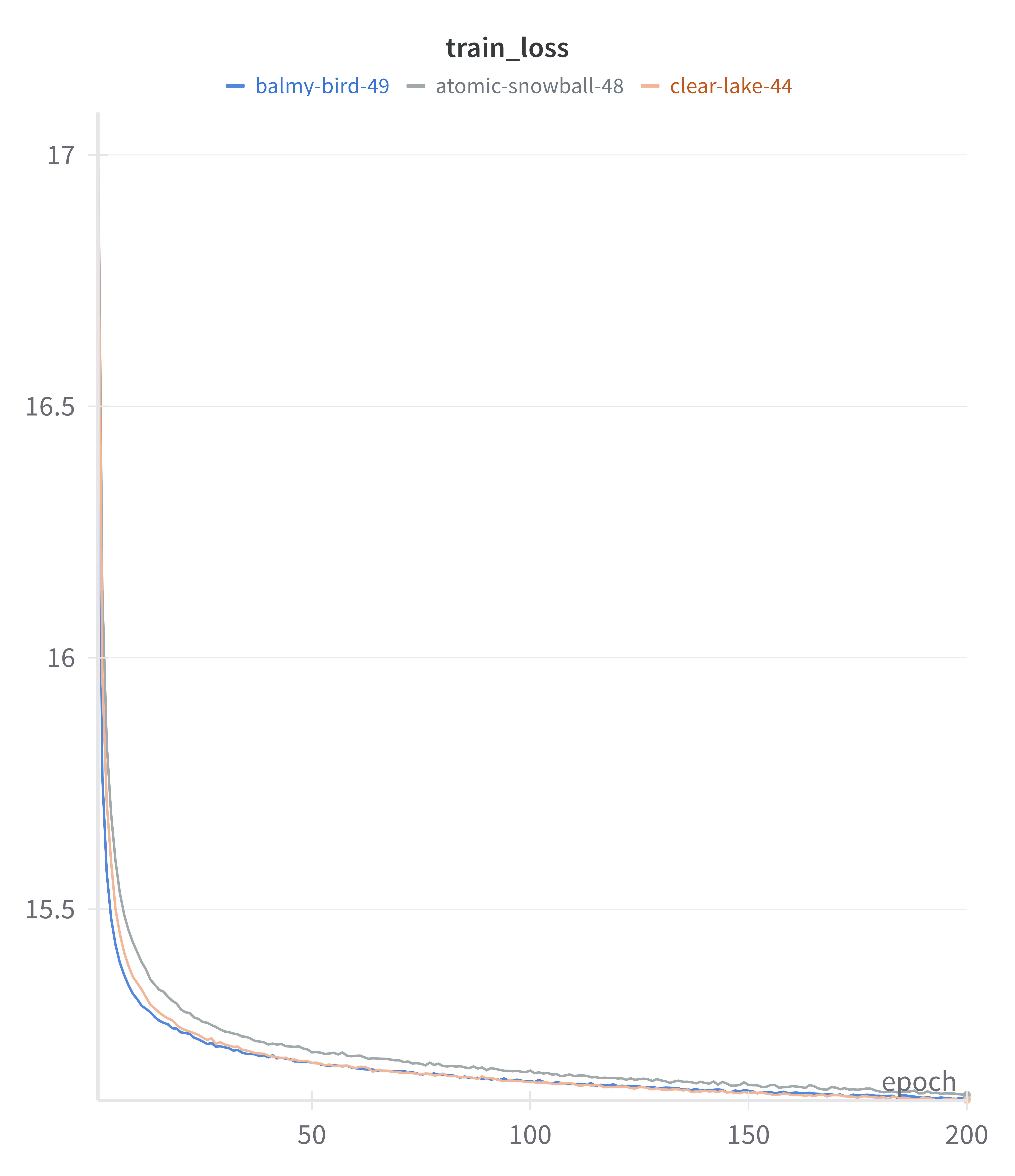}
			\caption{Training loss.}
			\label{fig:gnn-gat-train}
		\end{subfigure}
		\hfill
		\begin{subfigure}[b]{0.45\textwidth}
			\includegraphics[width=\textwidth]{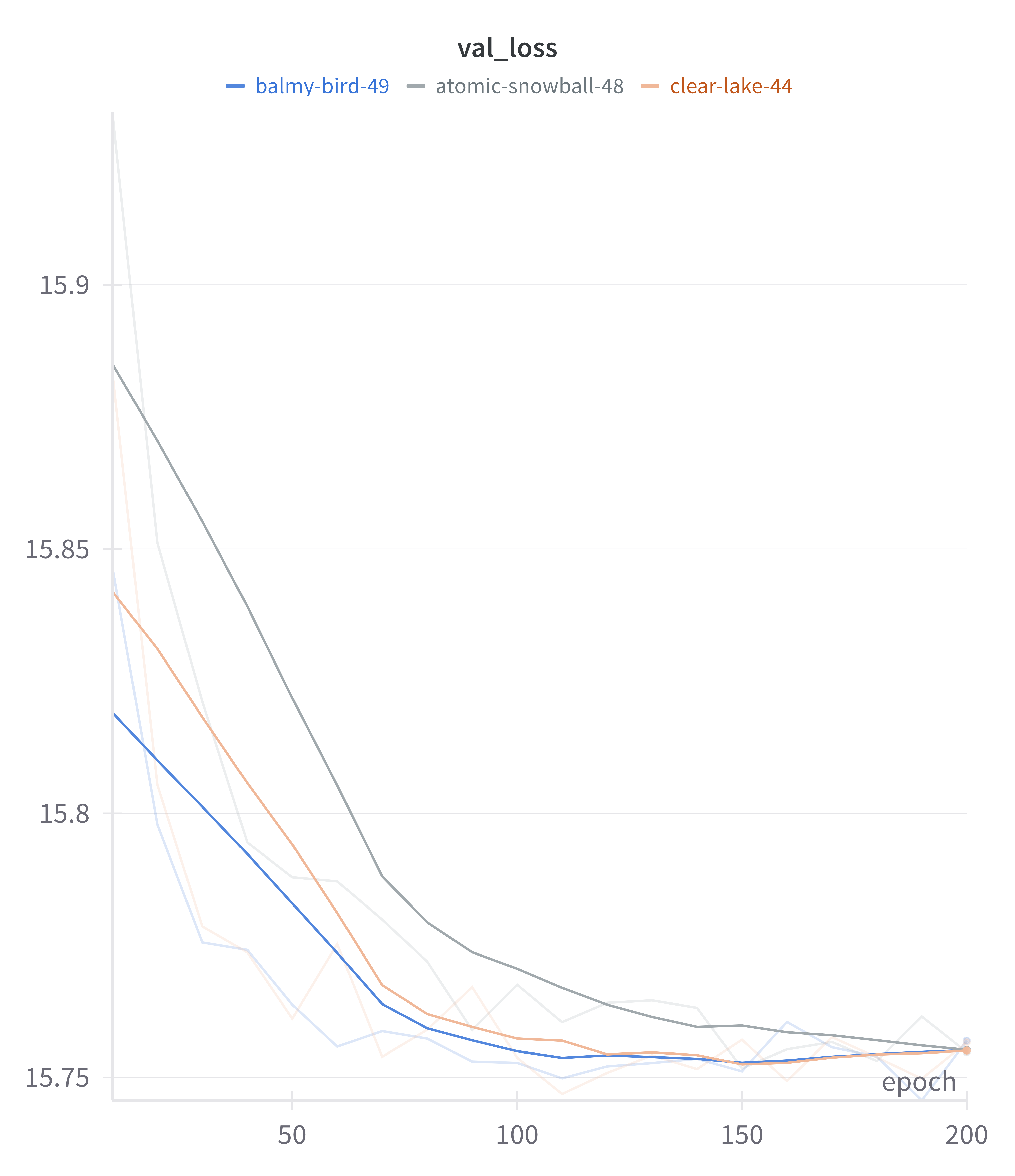}
			\caption{Validation loss.}
			\label{fig:gnn-gat-val}
		\end{subfigure}
	
		\caption{Training GNN-GATv2Conv encoder.}
		\label{fig:gnn-gat}
	\end{figure}

	We evaluated GNN‐GATv2Conv by fixing $f_{out}=64$ (since $f_{out}=128$ caused overfitting) and varying the number of convolutional layers: $n_{conv} \in \{1, 2, 3\}$. \Cref{table:gnn-gat} maps each run name to its $n_{conv}$ value. As shown in \cref{fig:gnn-gat}, all three configurations converged to nearly the same validation loss. We selected \textit{balmy‐bird‐49} as the best model because it reached convergence slightly faster and exhibited fewer fluctuations after epoch 50 compared to \textit{clear‐lake‐44}. We attribute this to oversmoothing: our graph is relatively small, so adding more GATv2 layers tends to blur node features, and a lower $n_{conv}$ yields optimal performance.

	\begin{table}
		\caption{Considered hyperparameters combinations for the GNN‐GATv2Conv.}
		\label{table:gnn-gat}
		\centering
		\begin{tabular}{ll}
			\toprule
			Run name            & $n_{conv}$ \\
			\midrule
			\textbf{balmy-bird-49} & \textbf{1} \\
			clear-lake-44       & 2         \\
			atomic-snowball-48  & 3         \\
			\bottomrule
		\end{tabular}
	\end{table}

	\paragraph*{GNN-SAGEConv and GNN-GATv2Conv Comparison}

\begin{figure}[htb]
		\centering
		\begin{subfigure}[b]{0.45\textwidth}
			\includegraphics[width=\textwidth]{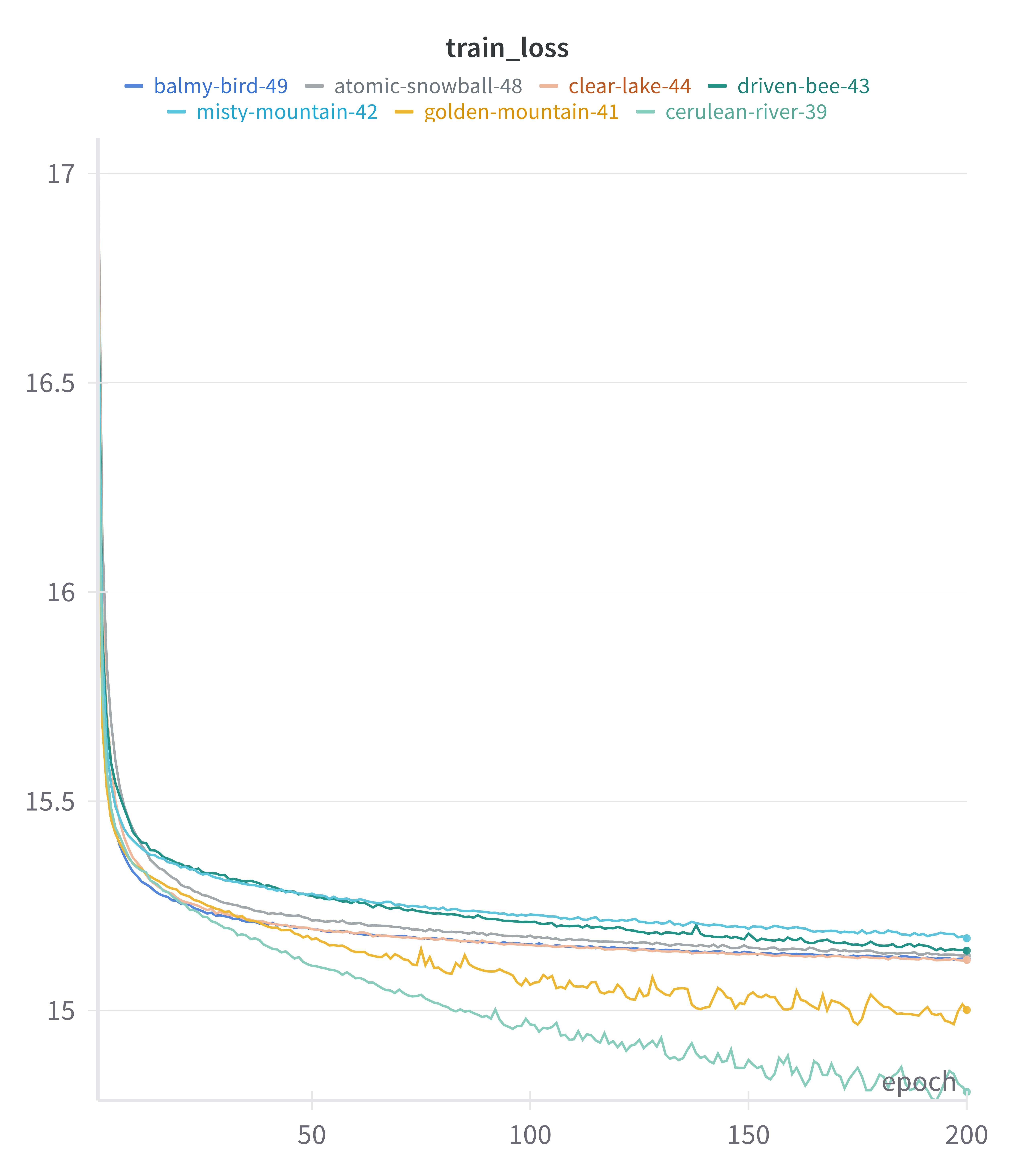}
			\caption{Training loss.}
			\label{fig:gnn-all-train}
		\end{subfigure}
		\hfill
		\begin{subfigure}[b]{0.45\textwidth}
			\includegraphics[width=\textwidth]{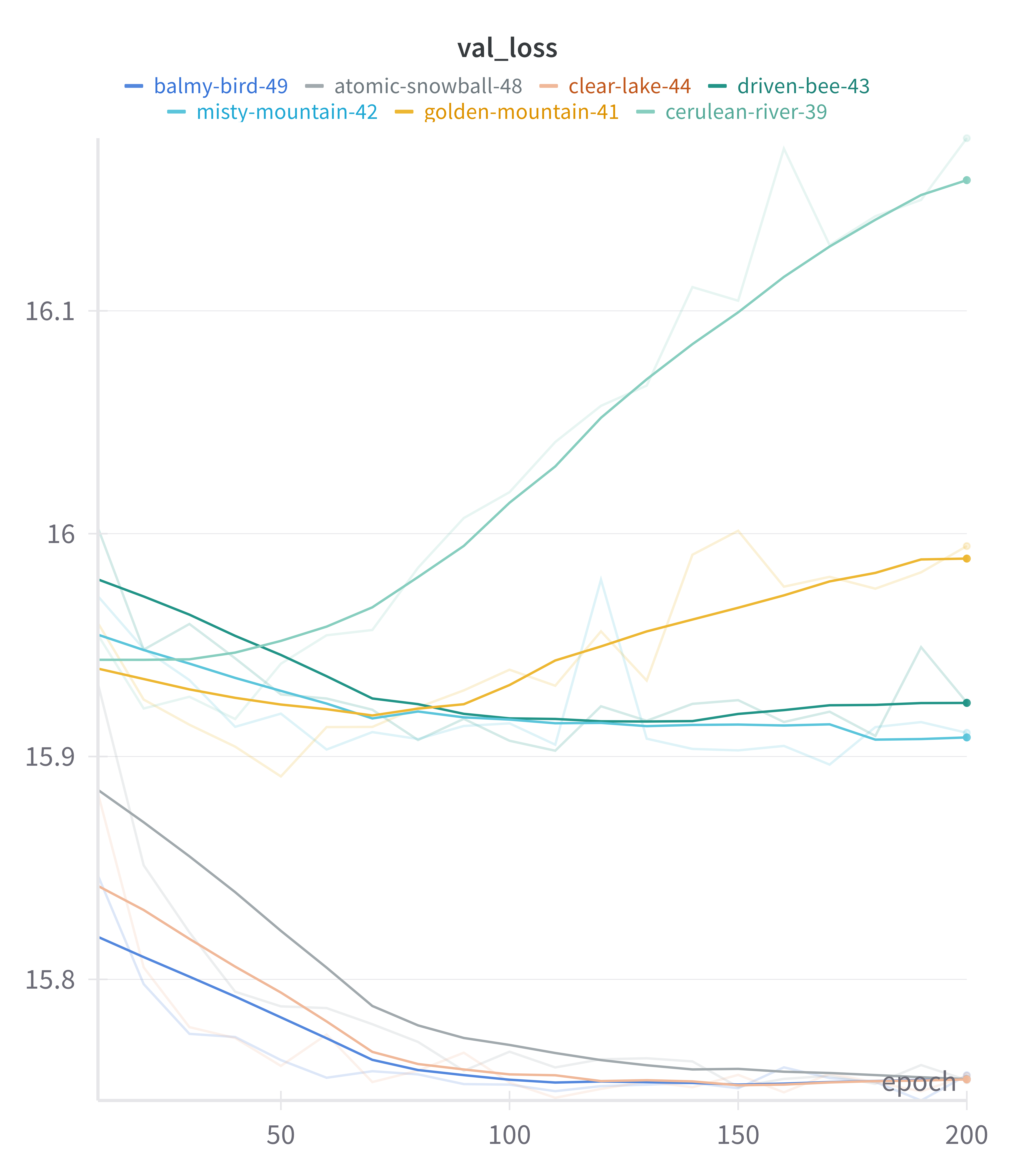}
			\caption{Validation loss.}
			\label{fig:gnn-all-val}
		\end{subfigure}
	
		\caption{Training GNN-SAGEConv and GNN-GATv2Conv based autoencoders.}
		\label{fig:gnn-all}
	\end{figure}

	Putting the results of both the GNN‐SAGEConv and the GNN‐GATv2Conv in \cref{fig:gnn-all}, we see that the validation loss curves for the GNN‐SAGEConv never overlap with those of the GNN‐GATv2Conv; all the GNN‐GATv2Conv curves lie below the GNN‐SAGEConv curves. Although the training losses are similar, the GNN‐GATv2Conv clearly outperforms the GNN‐SAGEConv on validation loss. This suggests that leveraging edge information in the GNN‐GATv2Conv is critical for the encoder–decoder task.

	\subsection{AE and GNN Encoders Comparison}
	
	We normalised the loss to compare the performance of the encoders. As the loss documented in the plots is the average trajectory loss, we have derived from the data that the average trajectory length in each data set (training, validation and test) is 13.3 and the median is 13. Therefore, dividing the observed loss by 13.3 gives the average loss per single step prediction. Each step consists of either 34 or 33 features, so dividing by this number we get the average loss per feature. Therefore the final normalization coefficient for the GNN encoder is $\frac{1}{13,3 \cdot 34}$, while for AE it is $\frac{1}{13,3\cdot 33}$.
	
	We normalized the loss to compare the performance of the encoders. Since the loss shown in the plots is the average trajectory loss, we found that the mean trajectory length in each dataset (training, validation, and test) is 13.3 (median 13). Dividing the observed loss by 13.3 yields the average loss per single‐step prediction. Each step consists of either 34 features (for GNN) or 33 features (for AE), so dividing by these numbers gives the average loss per feature. Thus, the normalization coefficient is $\frac{1}{13.3 \times 34}$ for the GNN encoder and  $\frac{1}{13.3 \times 33}$ for the AE.

	\begin{table}
		\caption{Final training loss and normalized loss for each autoencoder model.}
		\label{table:loss-comparison}
		\centering
		\begin{tabular}{lccc}
			\toprule
			Encoder–decoder    & Loss   & Normalized loss  & Absolute difference with AE \\
			\midrule
			AE                 & 15.44  & 0.035179         & 0                          \\
			GNN–SAGEConv       & 15.91  & 0.035184         & 0.000005                   \\
			GNN–GATv2Conv      & 15.75  & 0.034830         & 0.00034                    \\
			\bottomrule
		\end{tabular}
	\end{table}

	The normalization indicates that the best performing model is GNN-GATv2Conv, while GNN-SAGEConv and AE perform nearly the same and therefore share second place. The absolute difference in loss per feature between AE and GNN‐SAGEConv is marginal (5e-6), whereas the difference between AE and GNN‐GATv2Conv is about 3e-4 —substantially larger.

	\subsection{Action Injection Experiment}

	As an additional experiment, we considered an alternative autoencoders training approach without $a_t$ injection into the decoder input. We trained both GNN encoders with the same hyperparameters, $f_{out}=64$ and $n_{conv}=2$, for 200 epochs, performing validation every 10 epochs. The \textit{jumping‐water‐19} run corresponds to the architecture with action injection, while “upbeat‐star‐20” corresponds to the version without action injection.

\begin{figure}[htb]
		\centering
		\begin{subfigure}[b]{0.45\textwidth}
			\includegraphics[width=\textwidth]{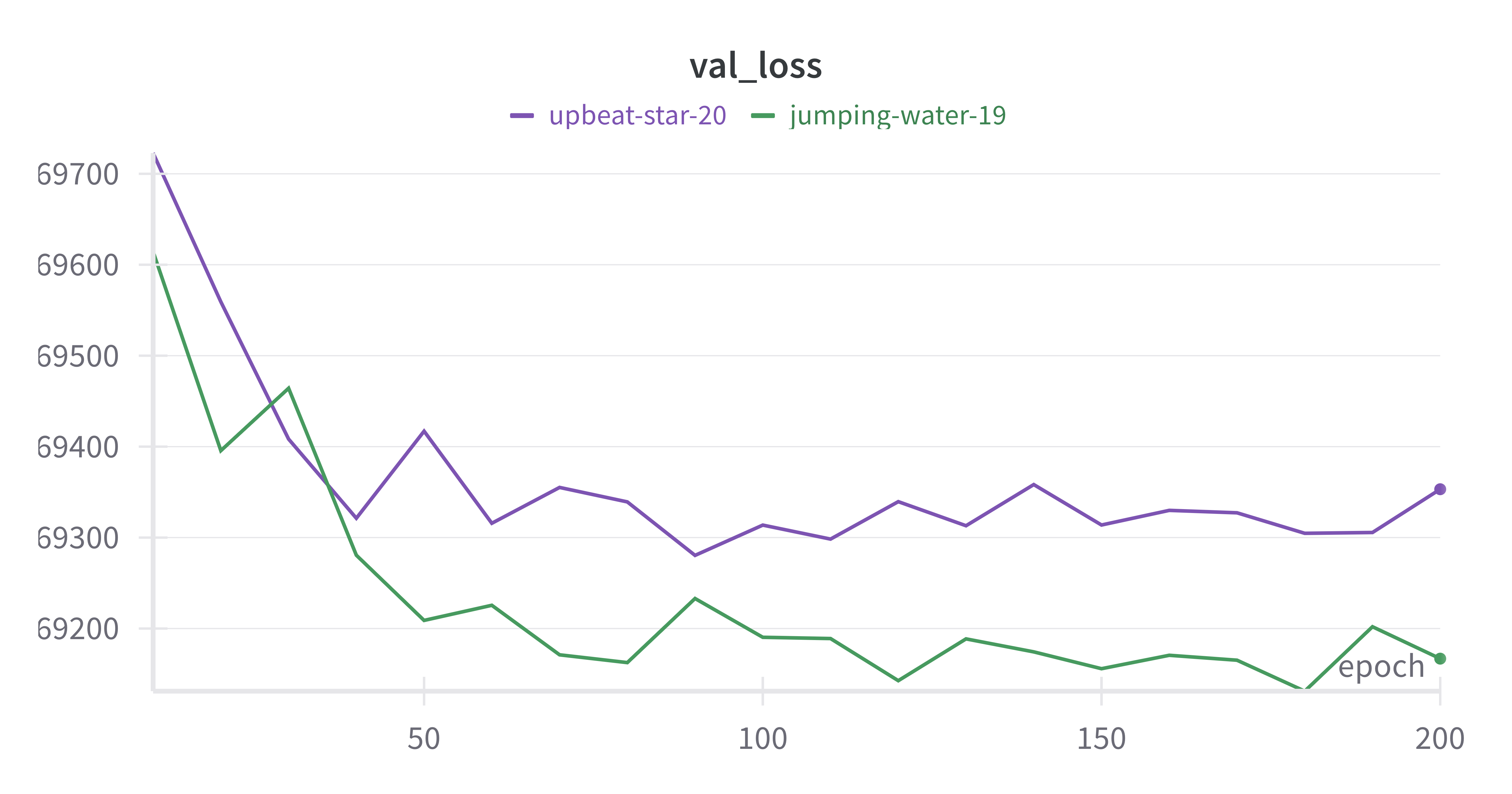}
			\caption{Validation loss.}
			\label{fig:inject_action_val_loss}
		\end{subfigure}
		\hfill
		\begin{subfigure}[b]{0.45\textwidth}
			\includegraphics[width=\textwidth]{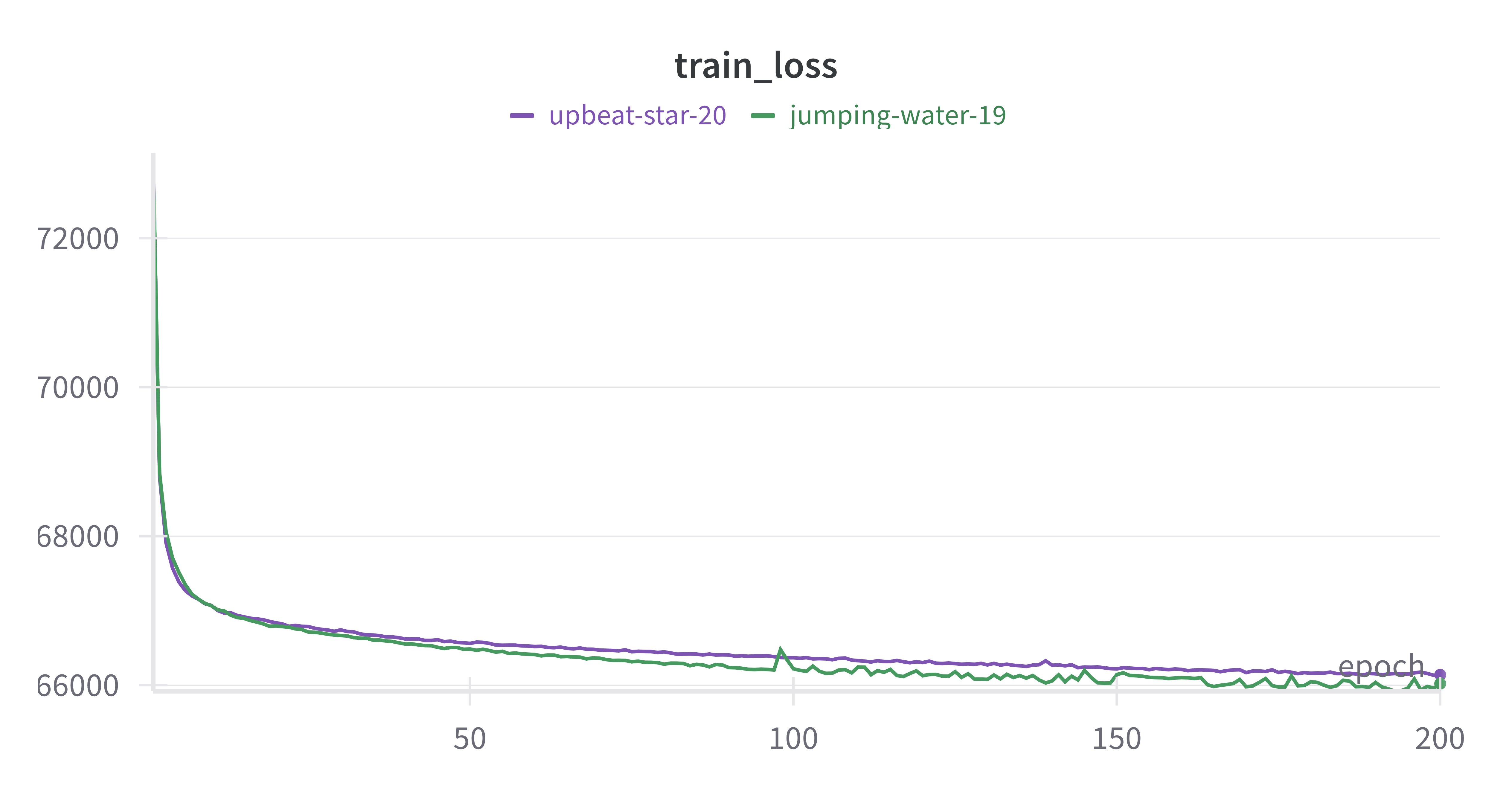}
			\caption{Train loss.}
			\label{fig:inject_action_train_loss}
		\end{subfigure}
		\caption{Experiments with and without action $a_t$ injection.}
		\label{fig:inject_action_experiment}
	\end{figure}

	In \cref{fig:inject_action_experiment}, the plotted losses are average batch losses. Each batch loss is the sum of all trajectory losses in that batch, and each trajectory loss is the sum of its per‐step losses. Although the training losses are similar, the validation loss for the architecture without action injection is noticeably higher. Therefore, in the final encoder–decoder design, we include action injection.

	\begin{figure}[htb]
		\centering
		\includegraphics[width=0.8\textwidth]{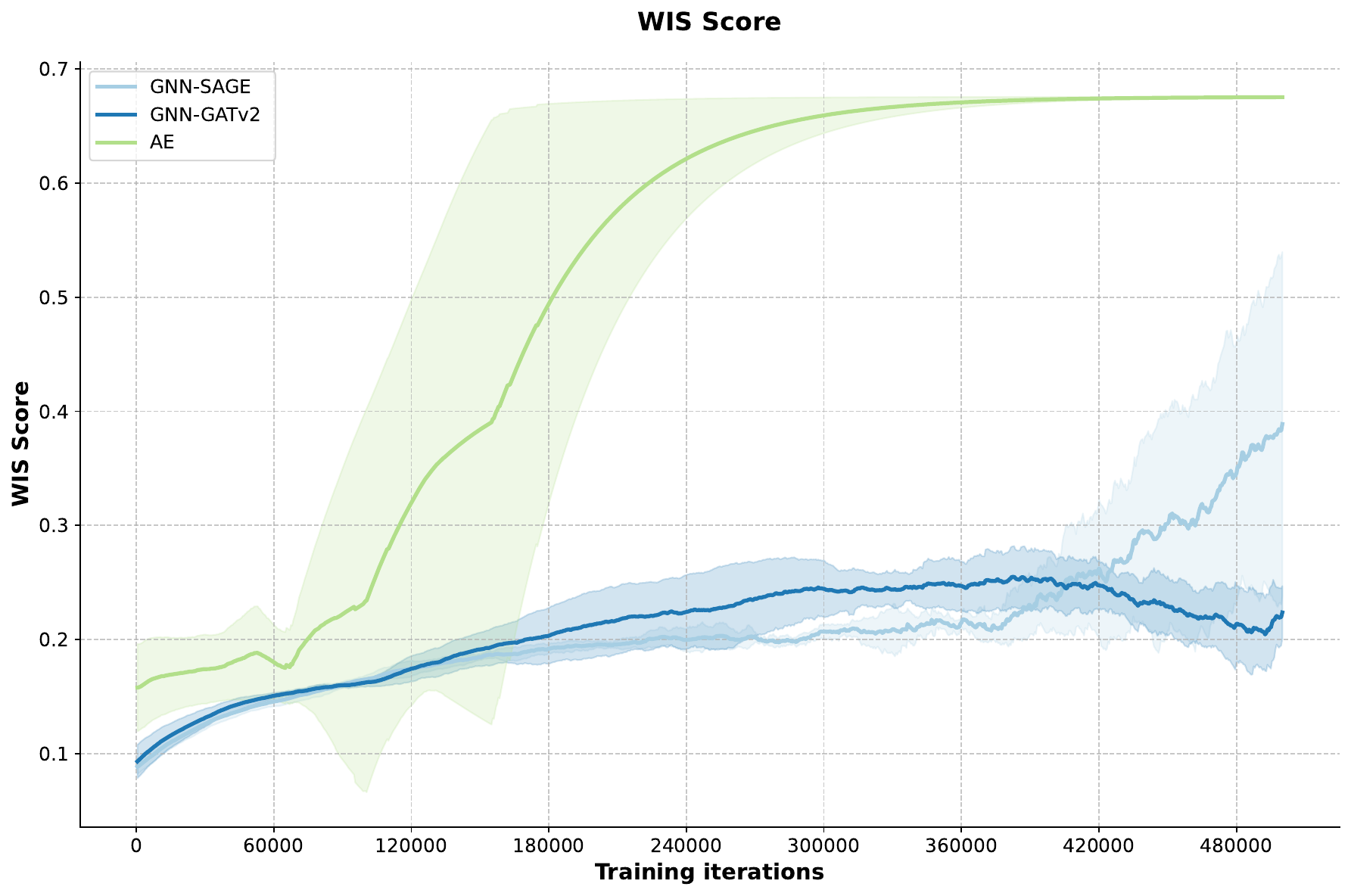}
	
		\caption{WIS on 500K training iteration on AE, GNN-SAGEConv and GNN-GATv2Conv.}
		\label{fig:wis-normal-5e5}
	
	\end{figure}

	\section{Policy Training with Untrained Encoders}
	
\begin{figure}[htb]
		\centering
		\includegraphics[width=0.8\textwidth]{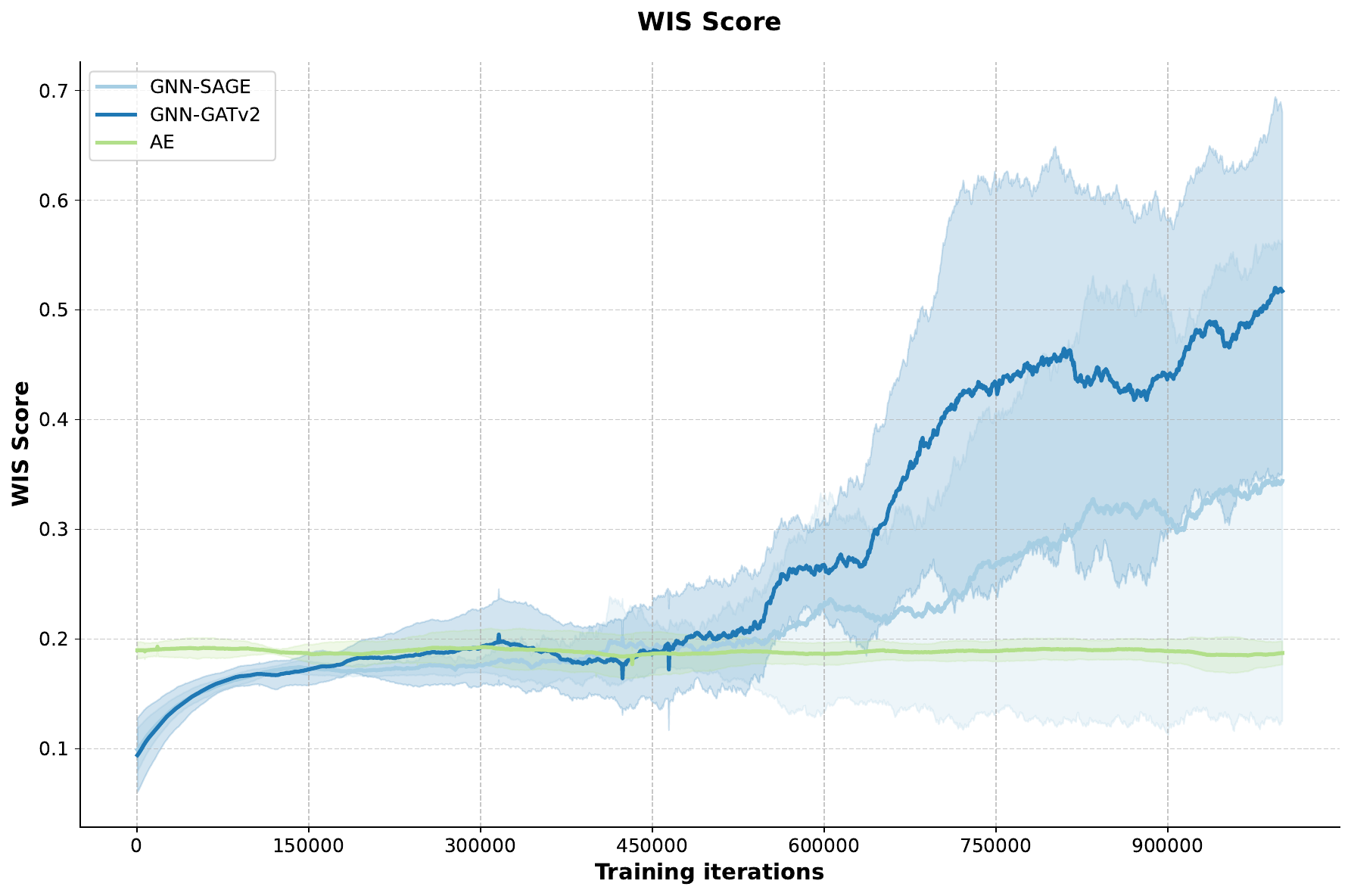}
	
		\caption{WIS on 1M training iterations with not trained encoders.}
		\label{fig:wis-no-train}
	
	\end{figure}

	We ran policy training four times for each encoder type (randomly initialized weights) using seeds 1234, 25, 32, and 2020 (see \cref{fig:wis-no-train}). Each run consisted of 1 million iterations, with evaluations every 500 iterations. The AE encoder failed to learn, while both GNN encoders—GATv2 and SAGEConv—achieved better policies than the AE. As expected, both GNNs exhibited high variance due to their untrained weights, with GATv2 outperforming SAGE.

	\begin{figure}[htb]
		\centering
		\begin{subfigure}[b]{0.45\textwidth}
			\centering
			\includegraphics[width=\textwidth]{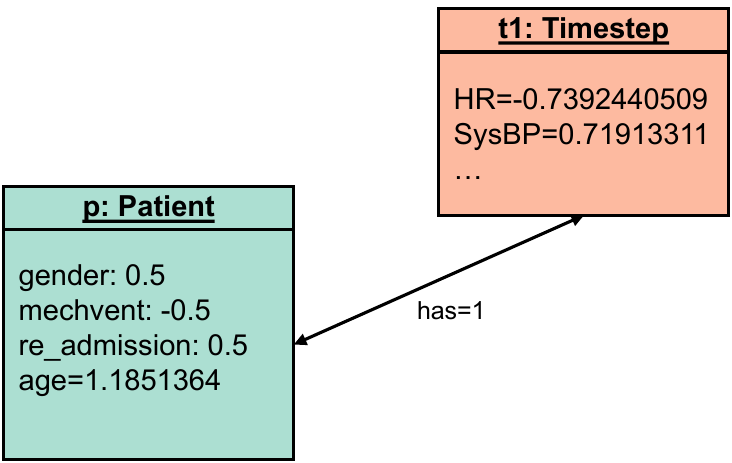}
			\caption{Snapshot at the first time step.}
			\label{fig:ts1}
		\end{subfigure}
		\hfill
		\begin{subfigure}[b]{0.45\textwidth}
			\centering
			\includegraphics[width=\textwidth]{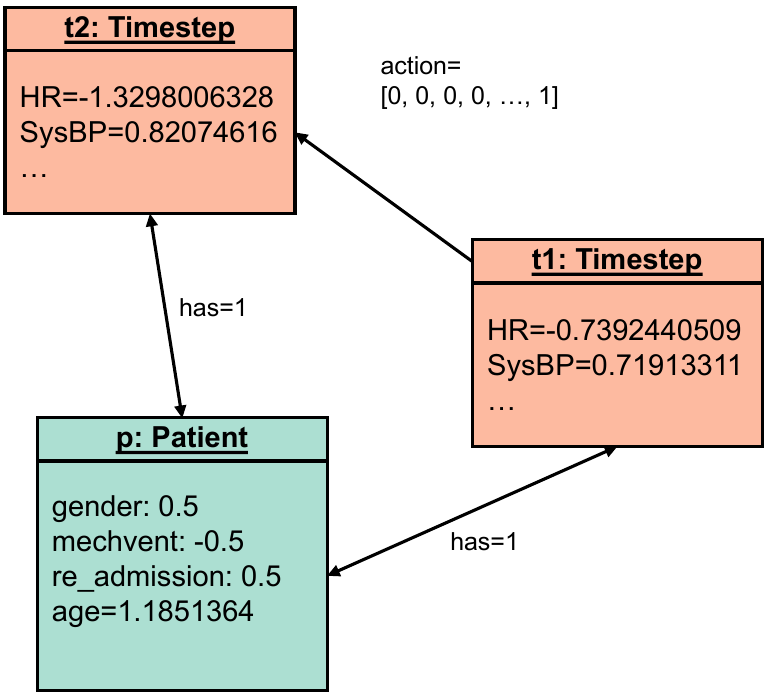}
			\caption{Snapshot at the second time step.}
			\label{fig:ts2}
		\end{subfigure}
	
		\vspace{1em}
	
		\begin{subfigure}[b]{0.45\textwidth}
			\centering
			\includegraphics[width=\textwidth]{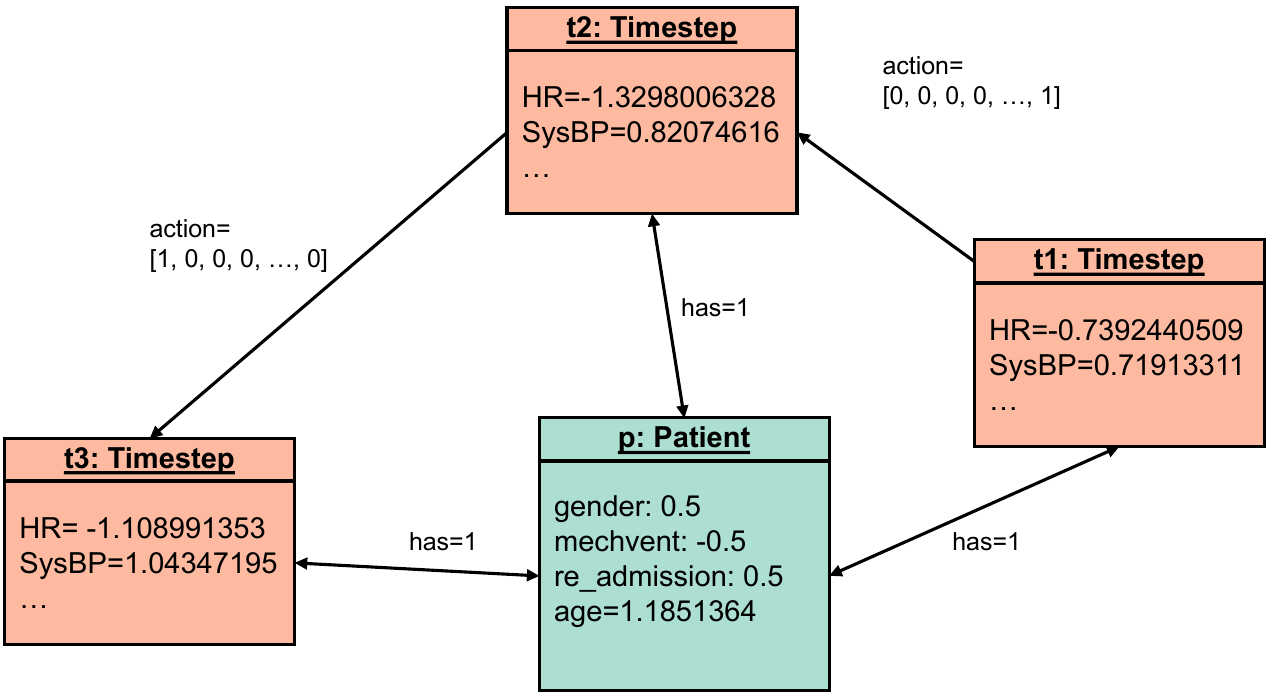}
			\caption{Snapshot at the third time step.}
			\label{fig:ts3}
		\end{subfigure}
		\hfill
		\begin{subfigure}[b]{0.45\textwidth}
			\centering
			\includegraphics[width=\textwidth]{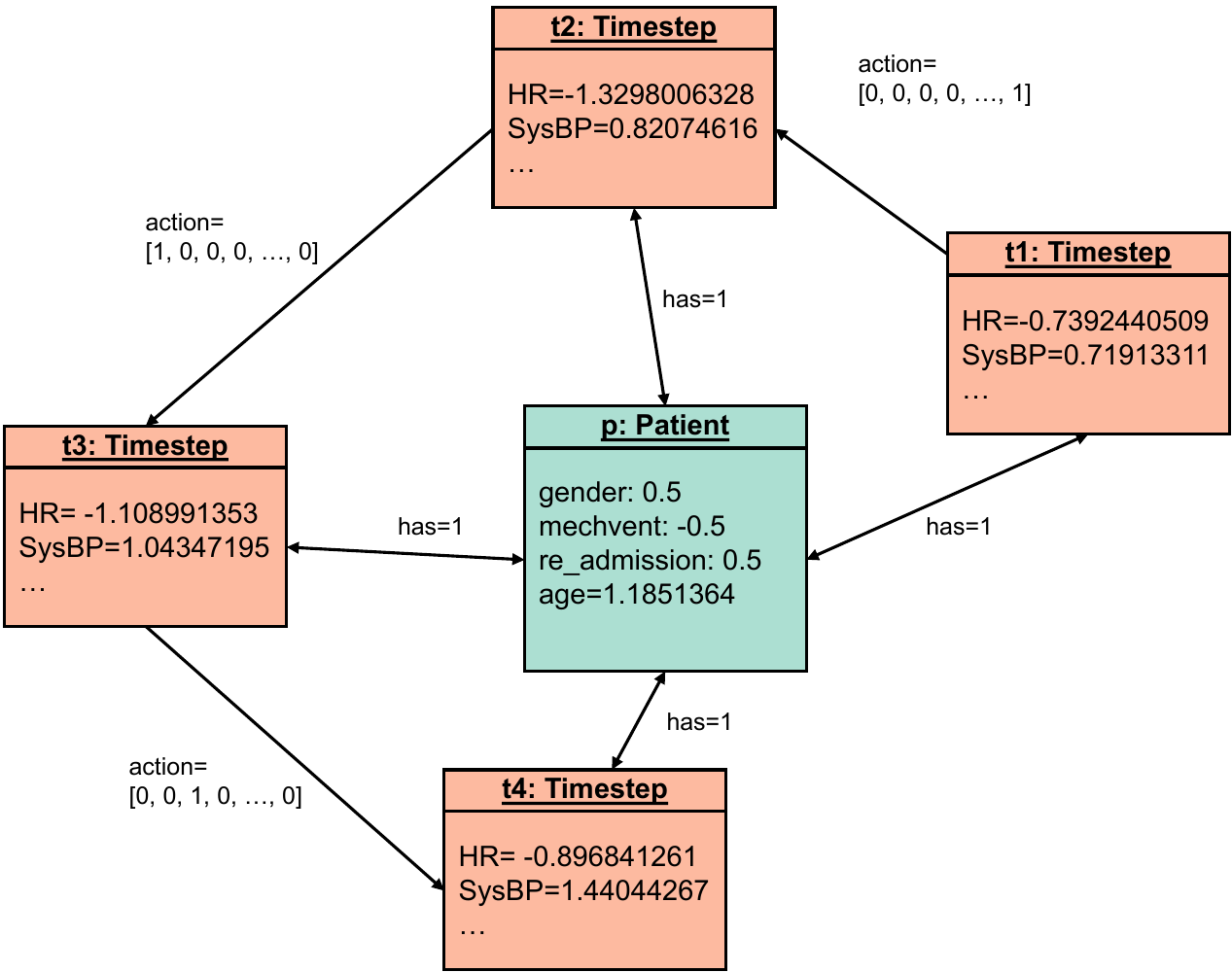}
			\caption{Snapshot at the fourth time step.}
			\label{fig:ts4}
		\end{subfigure}
	
		\caption{Graph evolution over the first four time steps.}
		\label{fig:graphsnapshots}
	\end{figure}

\begin{table}[htbp]
	\centering
	\caption{Patient features grouped into time-invariant ($\mathcal{I}$) and time-variant ($\mathcal{T}$) categories.}
	\renewcommand{\arraystretch}{1.3} 
	\resizebox{\textwidth}{!}{%
		\begin{tabular}{|l|l|p{9cm}|}
			\hline
			\textbf{\rule{0pt}{2.6ex}Category} & \textbf{Variable} & \textbf{Human-readable name\rule[-1.2ex]{0pt}{0pt}} \\ \hline
			\multirow{4}{*}{Time-invariant ($\mathcal{I}$)} 
			& o:gender        & Gender \\ 
			& o:age           & Age \\ 
			& o:re\_admission & ICU Re-admission \\ 
			& o:mechvent      & Mechanical Ventilation (yes/no at admission) \\ \hline
			\multirow{44}{*}{Time-variant ($\mathcal{T}$)} 
			& o:max\_dose\_vaso   & Maximum Dose of Vasopressor (over 4h) \\ 
			& o:Weight\_kg        & Weight (kg) \\ 
			& o:GCS               & Glasgow Coma Scale \\ 
			& o:HR                & Heart Rate \\ 
			& o:SysBP             & Systolic Blood Pressure \\ 
			& o:MeanBP            & Mean Blood Pressure \\ 
			& o:DiaBP             & Diastolic Blood Pressure \\ 
			& o:RR                & Respiratory Rate \\ 
			& o:Temp\_C           & Temperature (°C) \\ 
			& o:FiO2\_1           & Fraction of Inspired Oxygen (FiO$_2$) \\ 
			& o:Potassium         & Potassium \\ 
			& o:Sodium            & Sodium \\ 
			& o:Chloride          & Chloride \\ 
			& o:Glucose           & Glucose \\ 
			& o:Magnesium         & Magnesium \\ 
			& o:Calcium           & Calcium \\ 
			& o:Hb                & Hemoglobin \\ 
			& o:WBC\_count        & White Blood Cell Count \\ 
			& o:Platelets\_count  & Platelet Count \\ 
			& o:PTT               & Partial Thromboplastin Time \\ 
			& o:PT                & Prothrombin Time \\ 
			& o:Arterial\_pH      & Arterial Blood pH \\ 
			& o:paO2              & Partial Pressure of Oxygen (PaO$_2$) \\ 
			& o:paCO2             & Partial Pressure of Carbon Dioxide (PaCO$_2$) \\ 
			& o:Arterial\_BE      & Arterial Base Excess \\ 
			& o:HCO3              & Bicarbonate (HCO$_3^-$) \\ 
			& o:Arterial\_lactate & Arterial Lactate \\ 
			& o:SOFA              & Sequential Organ Failure Assessment (SOFA) Score \\ 
			& o:SIRS              & Systemic Inflammatory Response Syndrome (SIRS) Score \\ 
			& o:Shock\_Index      & Shock Index (HR / SBP) \\ 
			& o:PaO2\_FiO2        & PaO$_2$/FiO$_2$ Ratio \\ 
			& o:cumulated\_balance & Fluid Balance (cumulative) \\ 
			& o:SpO2              & Oxygen Saturation (SpO$_2$) \\ 
			& o:BUN               & Blood Urea Nitrogen \\ 
			& o:Creatinine        & Creatinine \\ 
			& o:SGOT              & Aspartate Aminotransferase (AST/SGOT) \\ 
			& o:SGPT              & Alanine Aminotransferase (ALT/SGPT) \\ 
			& o:Total\_bili       & Total Bilirubin \\ 
			& o:INR               & International Normalized Ratio (INR) \\ 
			& o:input\_total      & Fluid Input (total) \\ 
			& o:input\_4hourly    & Fluid Input (last 4h) \\ 
			& o:output\_total     & Fluid Output (total) \\ 
			& o:output\_4hourly   & Fluid Output (last 4h) \\ \hline
			
		\end{tabular}%
	}
	\label{table:all-features}
\end{table}

	\begin{table}
	\caption{Selected hyperparameters for the BCQ training.}
	\label{table:bcq-final-hyperparameters}
	\centering
	\begin{tabular}{llllll}
		\toprule
	 	BCQ threshold & Discount & Polyak target update & Target update frequency& Optimizer & Leaning rate\\
		\midrule
		 0.3 & 0.99 & True ($\tau=0,01$) & 1      & Adam & $1\mathrm{e}{-3}$   \\

		\bottomrule
	\end{tabular}
\end{table}

	\begin{table}
	\caption{Selected hyperparameters for the behavioral cloning training.}
	\label{table:bc-final-hyperparameters}
	\centering
	\begin{tabular}{llll}
		\toprule
		 Epochs& Optimizer  & Weight decay & Learning rate\\
		\midrule
		 5000 & Adam & 0.1  & $1\mathrm{e}{-4}$    \\

		\bottomrule
	\end{tabular}
\end{table}

\end{appendices}